%% file: deep_regression_3dv.tex
\documentclass[10pt,twocolumn,letterpaper,dvipsnames]{article}

\usepackage{3dv}
\usepackage{times}
\usepackage{epsfig}
\usepackage{graphicx}
\usepackage{amsmath}
\usepackage{amssymb}

\usepackage{microtype} 

\usepackage{bm}
\usepackage{textcomp} 
\usepackage{gensymb}
\usepackage{pifont}
\usepackage{array, makecell} %
\usepackage{xcolor} 
\newcommand{\cmark}{\textcolor{OliveGreen}{\ding{51}}}
\newcommand{\xmark}{\textcolor{Bittersweet}{\ding{55}}}

\usepackage{array}
\newcolumntype{K}[1]{>{\centering\arraybackslash}p{#1}}
\usepackage{graphbox} 
\usepackage{booktabs} 
\usepackage{multirow}
\usepackage{import}

\usepackage{xfp}
\NewExpandableDocumentCommand{\getlengthnumber}{O{pt}m}{%
	\fpeval{(#2)/(1#1)}%
}
\usepackage{listings}
\usepackage{stmaryrd} 

\DeclareMathOperator{\rank}{rank}
\DeclareMathOperator{\diag}{diag}

\DeclareMathOperator{\softmax}{softmax}
\DeclareMathOperator{\procrustes}{Procrustes}
\DeclareMathOperator{\SixD}{6D}

\usepackage{subfiles}

\usepackage[pagebackref=true,breaklinks=true,colorlinks,bookmarks=false]{hyperref}

\threedvfinalcopy 

\def\threedvPaperID{74} 

\ifthreedvfinal\fi

\begin{document}
	
\title{Deep Regression on Manifolds: A 3D Rotation Case Study}

\author{Romain Br\'egier\\
		NAVER LABS Europe\\
	{\tt\small romain.bregier@naverlabs.com}
}

\maketitle
\thispagestyle{empty} 

\begin{abstract}
Many machine learning problems involve regressing variables on a non-Euclidean manifold
-- \eg{} a discrete probability distribution, or the 6D pose of an object.
One way to tackle these problems through gradient-based learning is to use a differentiable function that maps arbitrary inputs of a Euclidean space onto the manifold.
In this paper, we establish a set of desirable properties for such mapping, and in particular highlight the importance of pre-images connectivity/convexity.
We illustrate these properties with a case study regarding 3D rotations.
Through theoretical considerations and methodological experiments on a variety of tasks, we review various differentiable mappings on the 3D rotation space, and conjecture about the importance of their local \emph{linearity}.
We show that a mapping based on Procrustes orthonormalization generally performs best among the mappings considered, but that a rotation vector representation might also be suitable when restricted to small angles.

\end{abstract}

\section{Introduction}
Deep neural networks typically produce feature vectors in a Euclidean space.
Euclidean spaces however are not suited to represent the output of many problems of computer vision, for which one might want to regress a point on a specific manifold.
Examples include \eg the prediction of a probability distribution for classification, or the regression of a 3D rotation matrix.
A strategy to circumvent this problem consists in using a differentiable function that maps feature vectors onto the manifold and enables to apply deep learning techniques onto this topological space (\eg{} using a \emph{softmax} for classification).
In this paper, we address this topic while focusing on the special use case of regressing 3D rotations.

\noindent\textbf{3D rotations}
3D rotation is an essential concept in many fields ranging from engineering to fundamental physics.
Regression of 3D rotations arises in many problems of computer vision,
such as absolute camera localization~\cite{kendall_posenet_2015, walch_image-based_2017},
object pose estimation~\cite{crivellaro_novel_2015, kehl_ssd-6d_2017, kanezaki_rotationnet_2018},
parameters identification for a rigid kinematic chain or for a human body model~\cite{omran_neural_2018},
or even as an intermediate step in a self-supervised depth estimation pipeline~\cite{zhou_unsupervised_2017}.

\noindent{\textbf{Regression on a manifold}}
The output of a deep learning architecture typically consists in a $n$-dimensional feature vector $\bm{x}$ of a Euclidean space $X=\mathbb{R}^n$. Mapping such arbitrary output to a target manifold $Y$ such as the 3D rotation space is not trivial, since $SO(3)$  is not homeomorphic to a Euclidean space.
Various approaches have therefore been explored to tackle regression on a target manifold:

\noindent\textit{-- Discretization:}
One solution consists in discretizing the target space, and reformulating the regression into a classification problem. Such approach has been used with success \eg in \cite{kanezaki_rotationnet_2018, kehl_ssd-6d_2017} for object attitude estimation. It may however not be satisfactory when precise regression is required, as the number of classes required is typically of the order of $(1/\alpha^d)$ with respect to a typical discretization step $\alpha$ and the dimension $d$ of the target manifold\footnote{assuming $Y$ to be a nonpathological manifold of finite measure.} ($d=3$ for 3D rotations).

\noindent\textit{-- Regression of an intermediary representation:}
An alternative consists in training the network to regress an intermediary representation using a surrogate loss, and using at test time a procedure to map this representation to an actual element of $Y$.
To estimate the pose of a rigid body from an image, 
Crivellaro~\etal{}~\cite{crivellaro_novel_2015} proposed to learn to regress a set of 2D heat maps representing projections of 3D keypoints, and to solve a PnP problem at test time.
He~\etal~\cite{he_pvn3d_2020} trained similarly a network to regress offsetted 3D keypoints used at test time to infer a pose, while others proposed to predict probability distributions on $SO(3)$ \cite{mohlin_probabilistic_2020, murphy_implicit-pdf_2021}. 

\noindent\textit{-- Differentiable mapping:}
A third alternative consists in including a differentiable function $f$ mapping from $X$ to the target space $Y$ as an ordinary layer of a deep learning architecture.
A model can then be trained either to minimize a loss function expressed on the target domain, or using $f(\bm{x})$ as an unsupervised intermediary representation~\cite{zhou_unsupervised_2017}.

In this work, we focus on the \emph{differentiable mapping} approach. Choice of a mapping function is nontrivial and can have a significant impact on downstream task performance.
A satisfactory explanation of what makes a good mapping for deep learning is still lacking, and we aim to help solving this issue through the following contributions:

\noindent 1) We establish in section~\ref{sec:mapping_manifold} a set of desirable properties for a mapping to enable gradient-based training and good generalization.

\noindent 2) We review in section~\ref{sec:deep_representations} existing mappings onto the 3D rotation space in the light of the aforementioned properties, and introduce \emph{RoMa}, a PyTorch library for 3D rotations.

\noindent 3) 
To assess the validity of our theoretical developments, we reproduce experiments of Zhou~\etal~\cite{zhou_continuity_2019} regarding regression of 3D rotations and extend them to other mappings and scenarios in section~\ref{sec:experiments}. We discuss in section~\ref{sec:discussion} lessons learned from this study in the particular case of 3D rotations.

\section{\label{sec:mapping_manifold}What makes a good differentiable mapping?}

To train a neural network to output representations on a $d$-dimensional manifold $Y$ through back-propagation, one can define a function $f: \bm{x} \in  X \rightarrow f(\bm{x}) \in Y$ to map an existing intermediary representation $x$ produced by the network and lying on a connected differentiable manifold $X$ (typically a feature vector of a Euclidean space $\mathbb{R}^n$) to an element $f(x)$ of $Y$.
One can then define an arbitrary loss function on the manifold, and perform training as usual, \eg through stochastic gradient-descent.
The choice of a mapping function $f$ has an important impact on actual test performance (see section~\ref{sec:experiments} and \cite{zhou_continuity_2019, levinson_analysis_2020} for 3D rotations), and understanding what makes a good mapping for deep learning is therefore of great practical interest.

\paragraph{Prior work}
The scope of traditional geometric deep learning literature focuses on processing data defined on non-Euclidean domain or on characterizing its underlying structure~\cite{bronstein_geometric_2017, cao_comprehensive_2020}, and therefore differs from ours.
Zhou \etal~\cite{zhou_continuity_2019} recently considered the problem of mapping representations between different spaces and stated that a mapping $f$ should be surjective and satisfy a notion of ``\emph{continuity}'', consisting in the existence of a continuous right inverse $g: Y \rightarrow X$.
While such property seems desirable, it consists in a very loose criterion.
As an example, it is satisfied by the following function from the set of $3 \times 3$ matrices to the 3D rotation space:
$
\bm{M} \in \mathcal{M}_{3,3}(\mathbb{R}) \rightarrow
[\bm{M} \text{ if } \bm{M} \in SO(3), \bm{I} \text{ else}]
$.
Such mapping is unlikely to be useful for deep learning however, as it maps almost any input to the identity rotation and admits null derivatives almost everywhere.

We therefore go further in the analysis, and try in this section to characterize desirable properties for a mapping to allow proper gradient-based machine learning.

\subsection{Desirable properties}
For a mapping to be suitable to a deep learning application, a minimal set of properties have first to be satisfied:

\noindent\textbf{$Y$ should be a connected differentiable manifold.}
Differentiability is indeed required to apply back-propagation during training, and one could not hope reaching an arbitrary target $y \in Y$ from an arbitrary starting point through small gradient-descent displacements if the target space were not composed of a single connected component.
A finite set of connected components could be considered nonetheless by casting regression into a classification problem associated with regression on each connected parts.

\noindent\textbf{$f$ should be surjective, \ie $f(X)=Y$.}
Satisfying this condition is indeed required to be able to predict any arbitrary output $y \in Y$.

\noindent\textbf{$f$ should be differentiable.}
This property is required to allow gradient-based optimization, which is the foundation of deep learning. It implies that $f$ should be continuous.
\\[-1em] 

While not strictly required, additional properties can further help training and generalization:

\noindent\textbf{Full rank Jacobian: the Jacobian of $f$ should be of rank $d$ (the dimension of $Y$) everywhere to ensure training capabilities.}
This property ensures that one can always find an infinitesimal displacement to apply to $x$ in order to achieve an arbitrary infinitesimal displacement of the output $f(\bm{x)}$, therefore that there is always a signal to back-propagate during training.
In particular, it guarantees convergence of gradient descent towards a global minimum of $x \rightarrow \mathcal{L}(f(x))$, when trained with a convex differentiable loss $\mathcal{L}$ admitting a lower bound.
Such guarantee may not be essential for mappings onto a high dimensional space where one might find a direction to diminish the loss in practice (\eg ReLU activation is a $\mathbb{R}^n \rightarrow (\mathbb{R}^+)^n$ mapping that does not satisfy this property), but it is important in the low dimensional regime (\eg $d=3$ in the case of 3D rotations).

\begin{figure*}
	\centering
	\footnotesize
	{
		\newlength{\myfigwidth}
		\setlength{\myfigwidth}{0.8in}
		\newlength{\firstcolwidth}		
		\setlength{\firstcolwidth}{0.98in}
		\newlength{\mycolwidth}
		\setlength{\mycolwidth}{0.95in}
		\setlength{\tabcolsep}{1pt}
		\renewcommand{\arraystretch}{1.0}
		\noindent
		\begin{tabular}{>{\centering\arraybackslash}m{\firstcolwidth}K{\mycolwidth}K{\mycolwidth}K{\mycolwidth}K{\mycolwidth}K{\mycolwidth}K{\mycolwidth}}
			\cmidrule[1.5pt]{1-6}
			Mapping function $f$
			& \includegraphics[align=c,width=\myfigwidth]{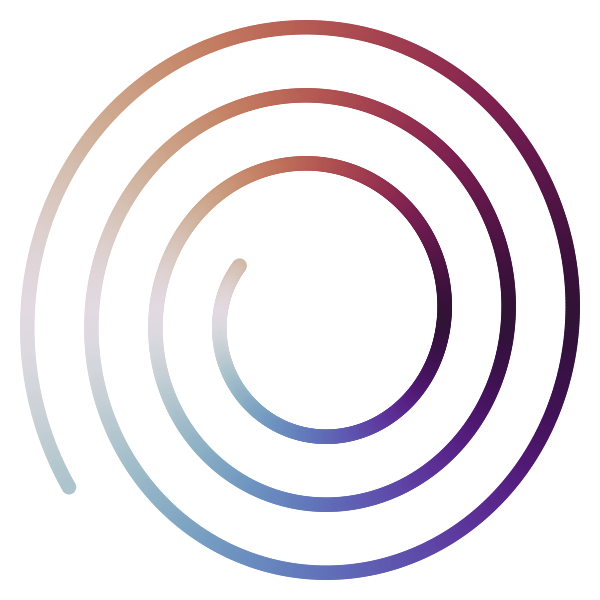} 
			& \includegraphics[align=c,width=\myfigwidth]{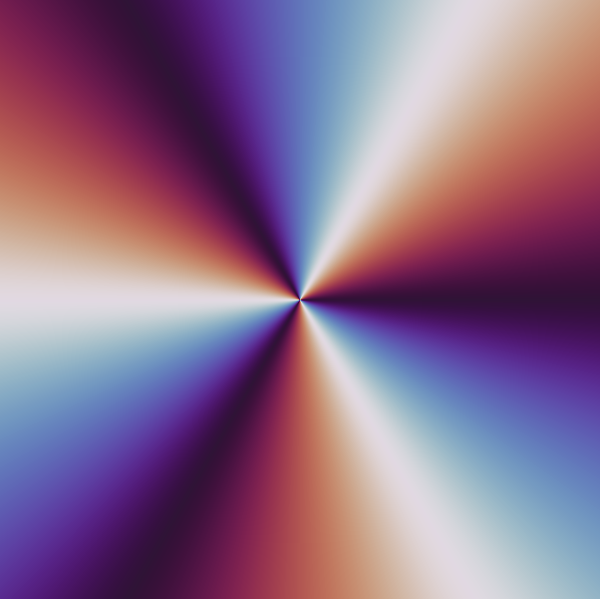}
			& \includegraphics[align=c,width=\myfigwidth]{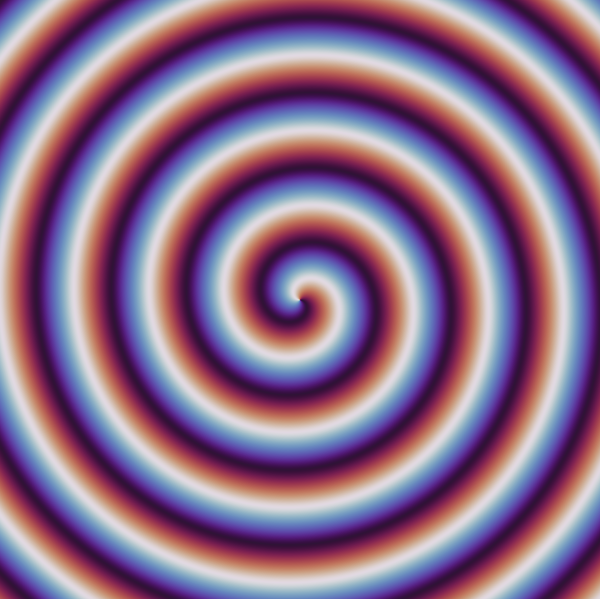}
			& \includegraphics[align=c,width=\myfigwidth]{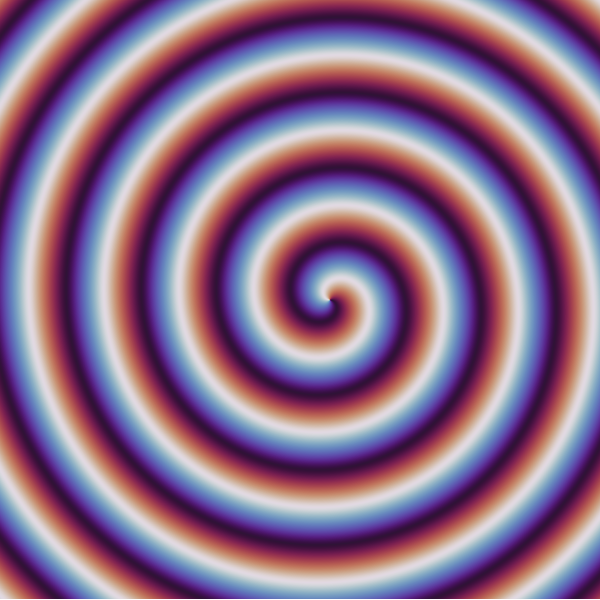}
			& \includegraphics[align=c,width=\myfigwidth]{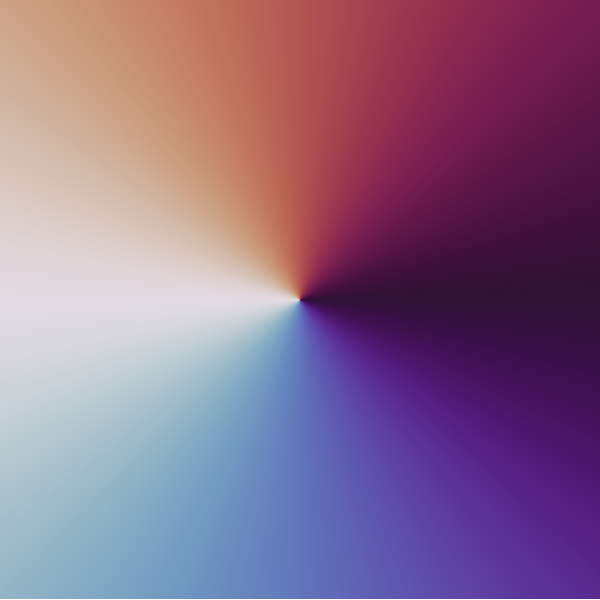}
			&
			\multirow{3}{*}{\parbox{\mycolwidth}{\centering Network architecture: \\ 		 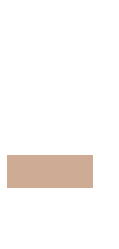}}
			\\
			\\[-8pt]
			
			& \parbox[c]{\mycolwidth}{\centering(a) \\ $x \in \mathbb{R} \rightarrow e^{i x}$}
			& \parbox[c]{\mycolwidth}{\centering(b) $x \in \mathbb{C} \rightarrow e^{3 i \arg(x)}$}
			& \parbox[c]{\mycolwidth}{\centering(c) $x \in \mathbb{C} \rightarrow e^{i (\arg(x) + 5 |x|)}$}
			& \parbox[c]{\mycolwidth}{\centering(d) $x \in \mathbb{C} \rightarrow {e^{i (\arg(x - o) + 5 |x - o|)}}$}
			& \parbox[c]{\mycolwidth}{\centering(e) \\ $x \in \mathbb{C} \rightarrow  x/ |x|$}
			&
			\\
			
			\cline{1-6}
			Connected pre-images
			&
			\xmark
			& \xmark
			& \cmark
			& \cmark
			& \cmark
			&
			\\
			
			Convex pre-images
			&
			\xmark
			& \xmark
			& \xmark
			& \xmark
			& \cmark
			&
			\\
			
			\cline{1-6}
			\\[-8pt] 
			\parbox[c]{\firstcolwidth}{\centering Training set \\ representations $x$}
			& \includegraphics[align=c,width=\myfigwidth]{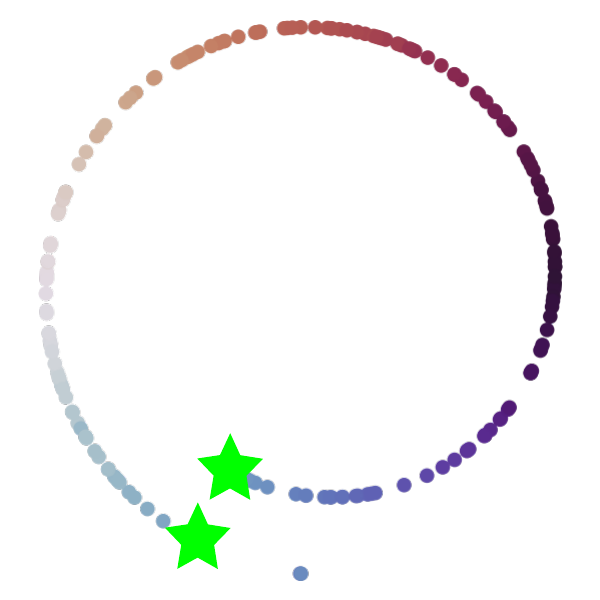}
			& \includegraphics[align=c,width=\myfigwidth]{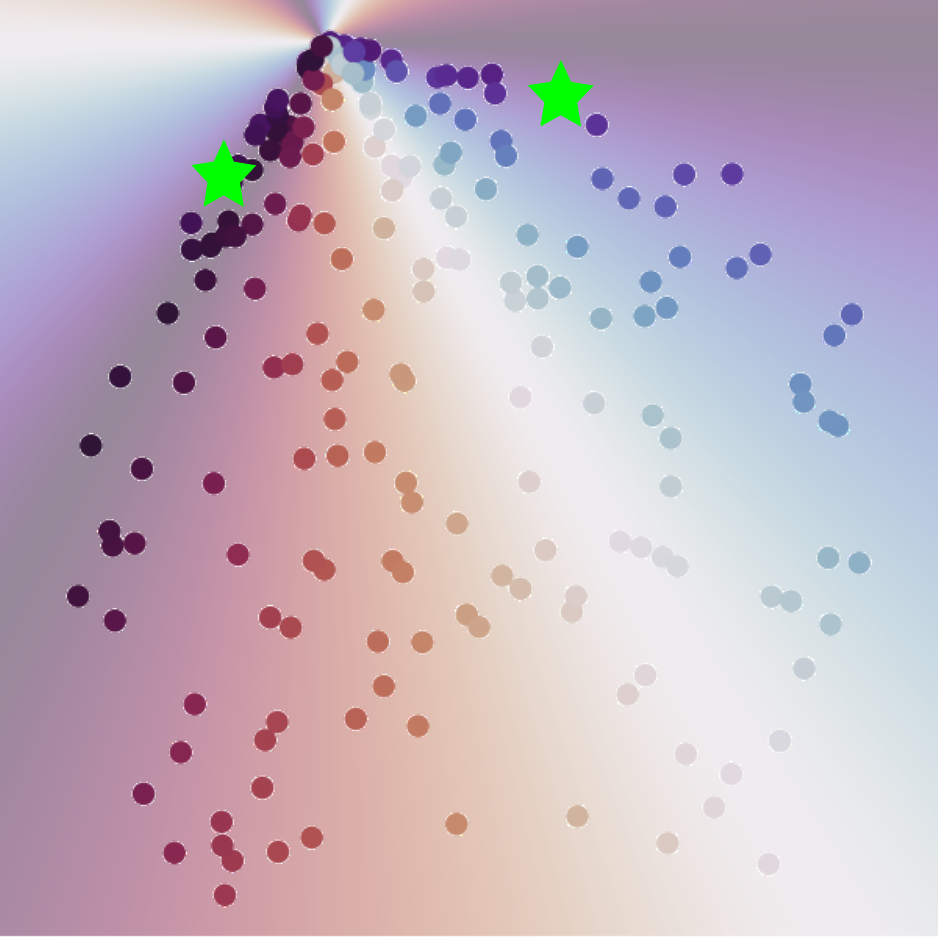}
			& \includegraphics[align=c,width=\myfigwidth]{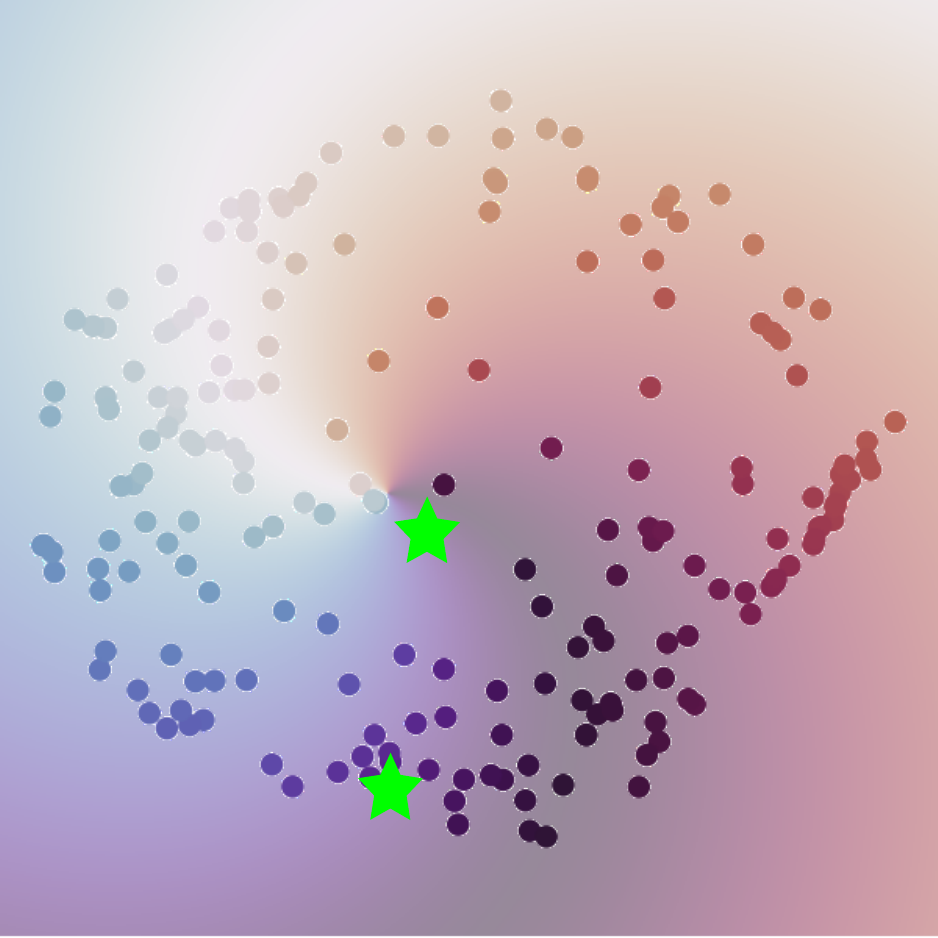}
			& \includegraphics[align=c,width=\myfigwidth]{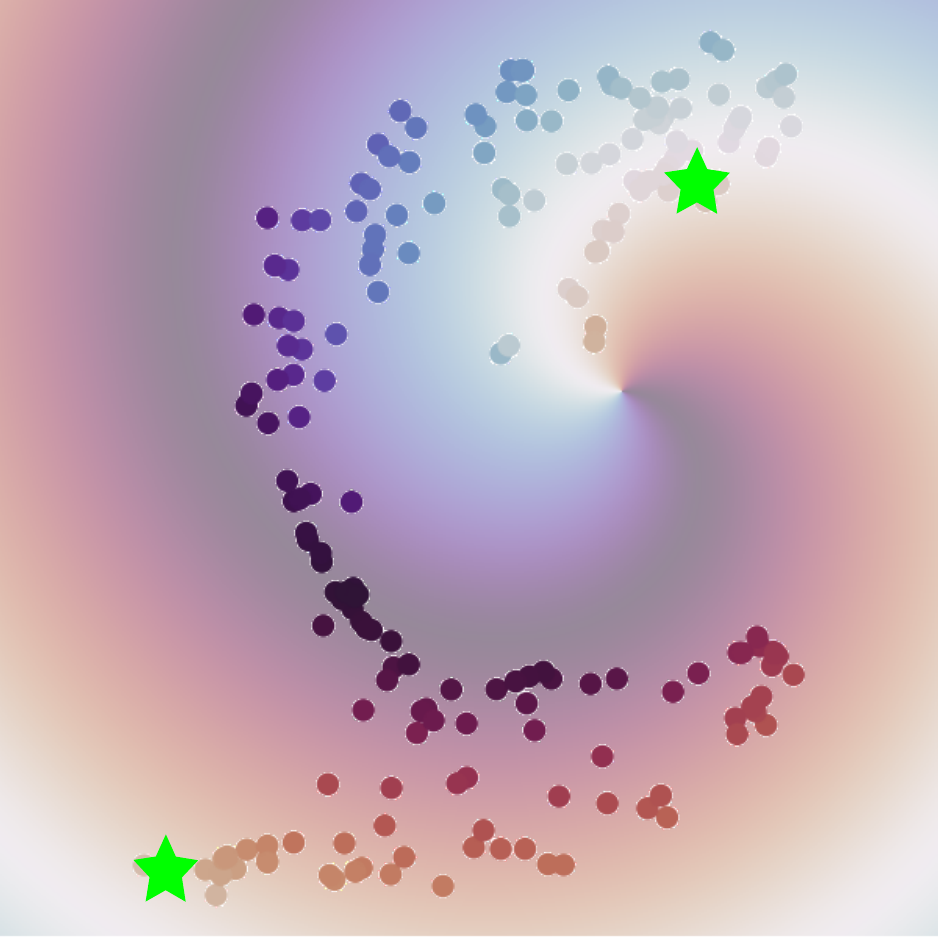}
			& \includegraphics[align=c,width=\myfigwidth]{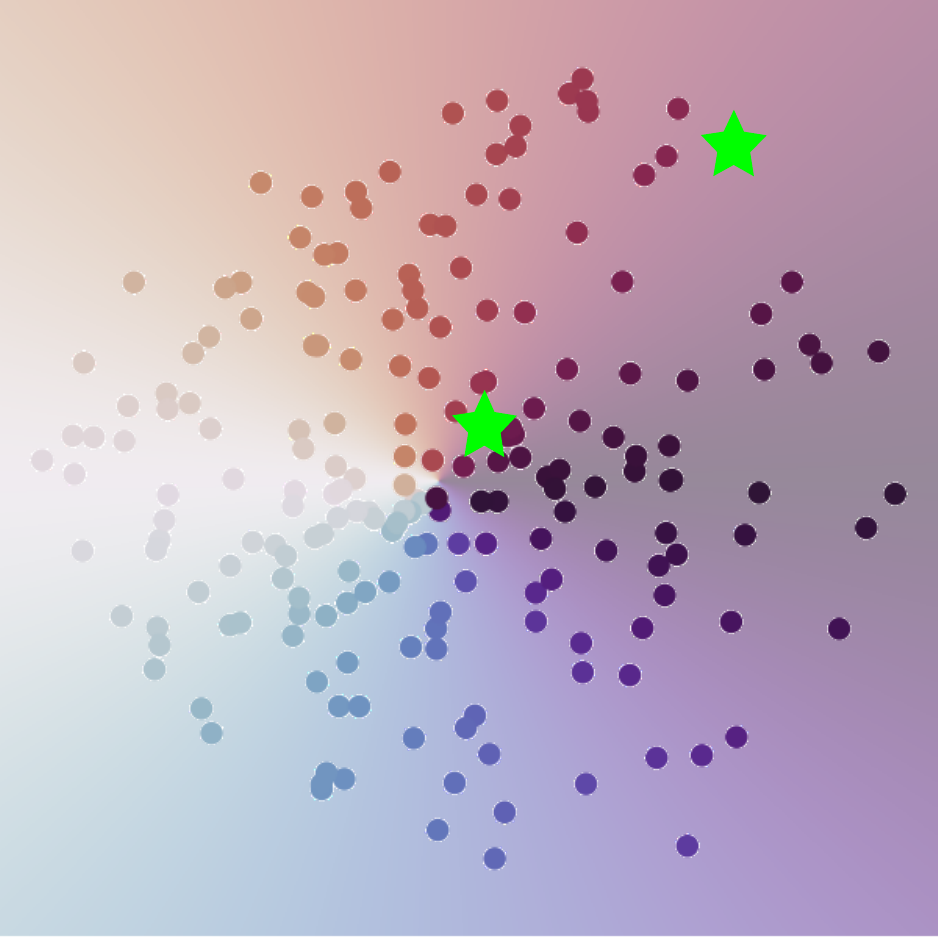}
			& 
			\\
			
			\\[-9pt] 
			\parbox[c]{\firstcolwidth}{\centering Test set \\ representations $x$}
			& \includegraphics[align=c,width=\myfigwidth]{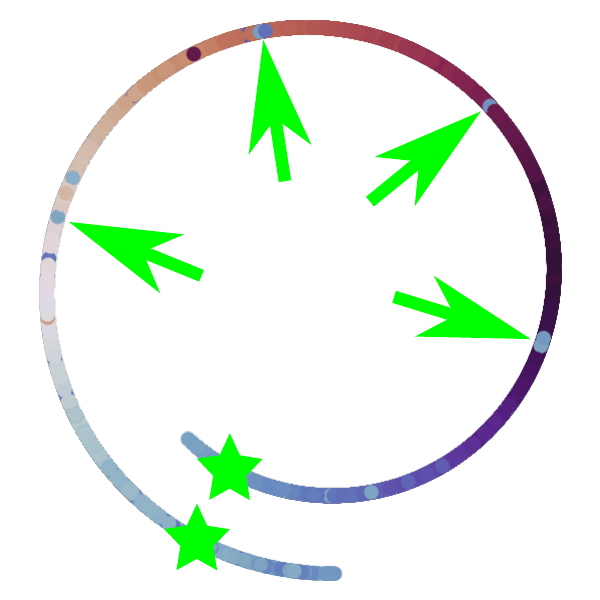}
			& \includegraphics[align=c,width=\myfigwidth]{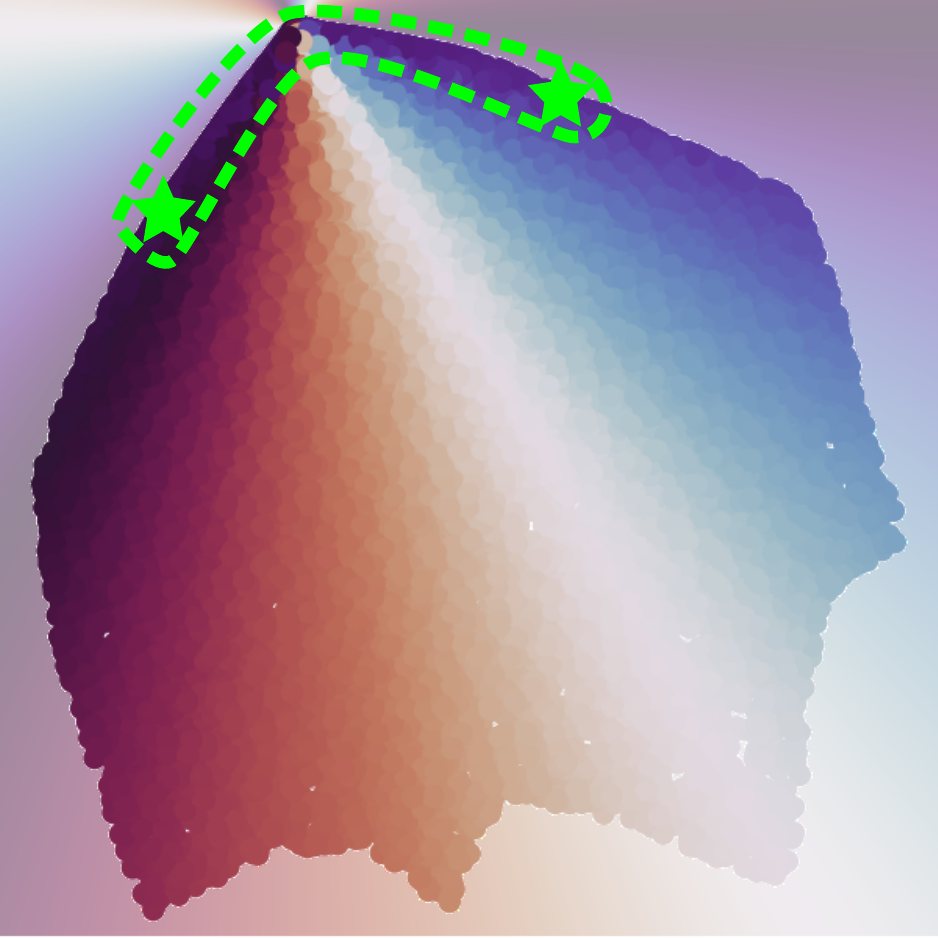}
			& \includegraphics[align=c,width=\myfigwidth]{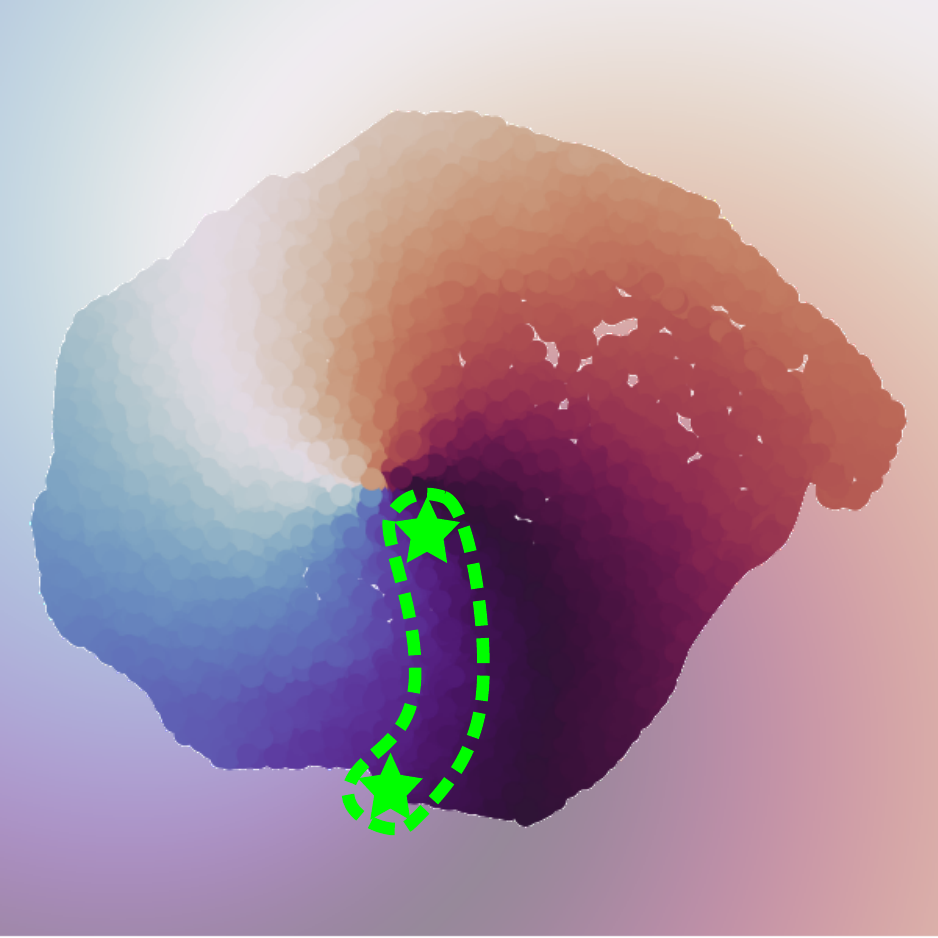}
			& \includegraphics[align=c,width=\myfigwidth]{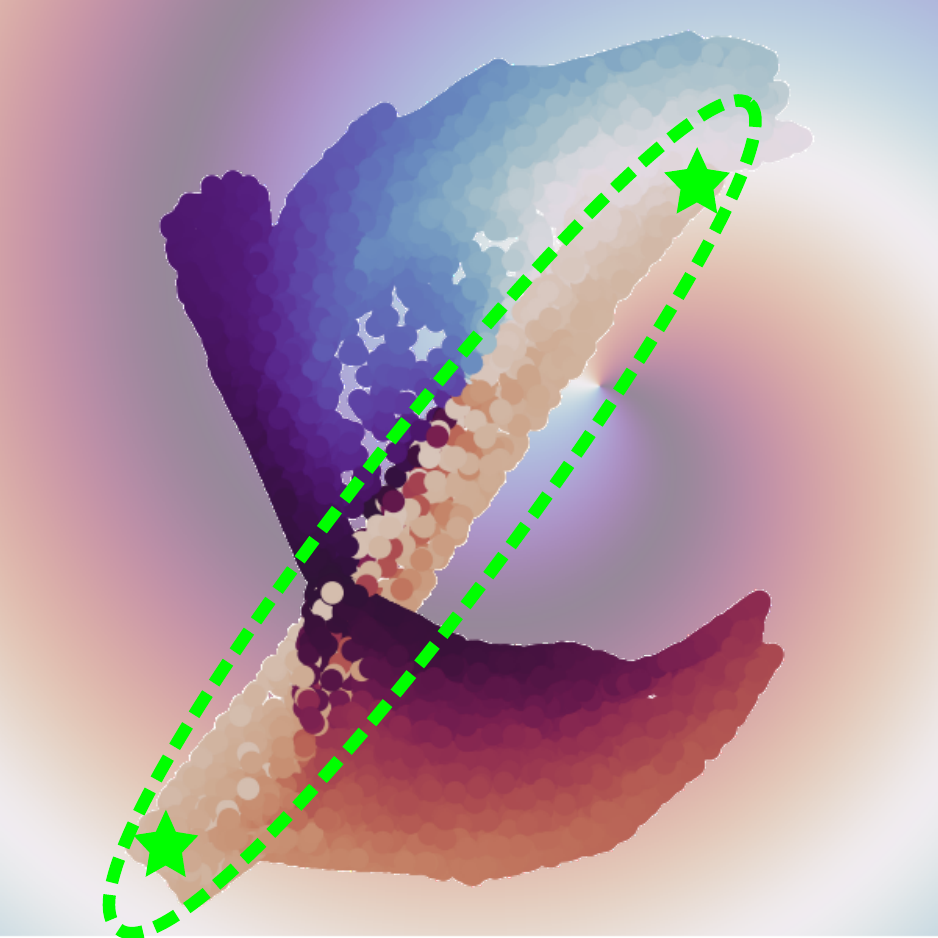}
			& \includegraphics[align=c,width=\myfigwidth]{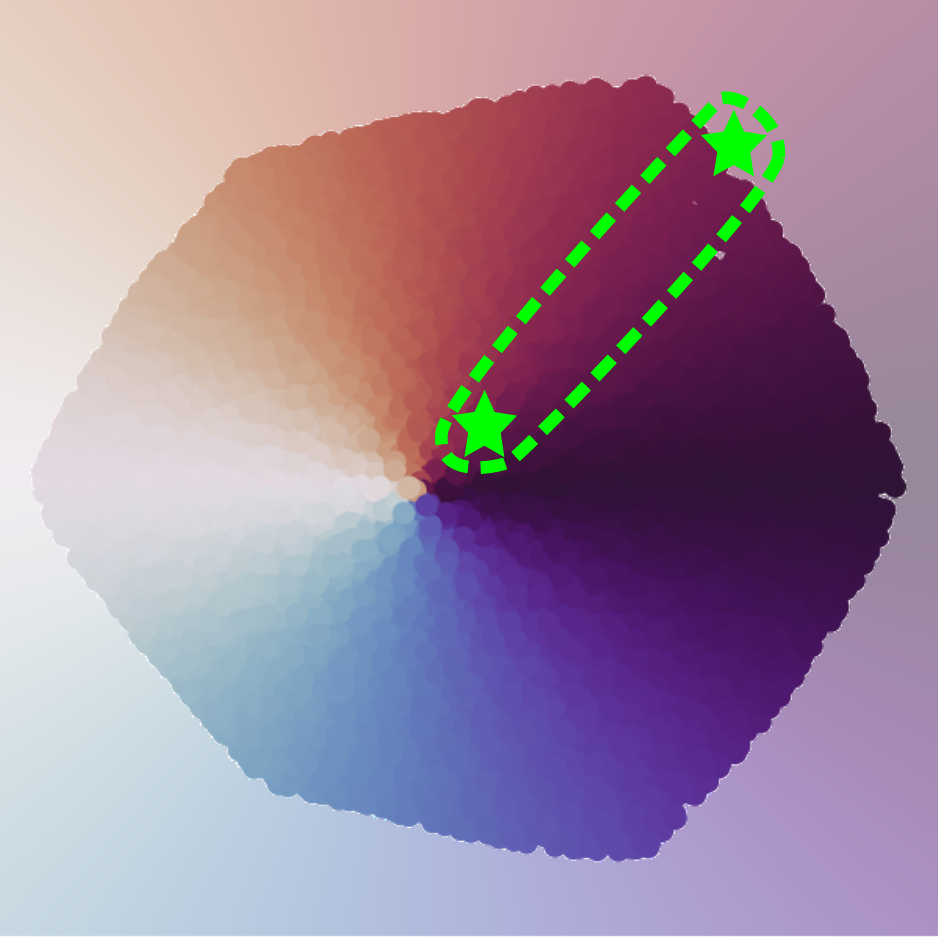}
			&
			\parbox{\mycolwidth}{\centering \vspace{50pt} Target predictions$^\dagger$:}
			\\
			
			\\[-9pt] 
			\cline{1-6}
			\parbox[c]{\firstcolwidth}{\centering{} Test set \\ hue predictions$^\dagger$}
			& \includegraphics[align=c,width=\myfigwidth]{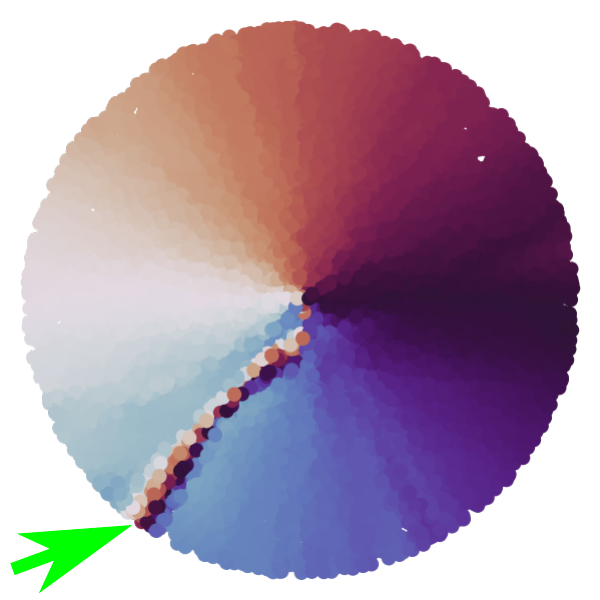}
			& \includegraphics[align=c,width=\myfigwidth]{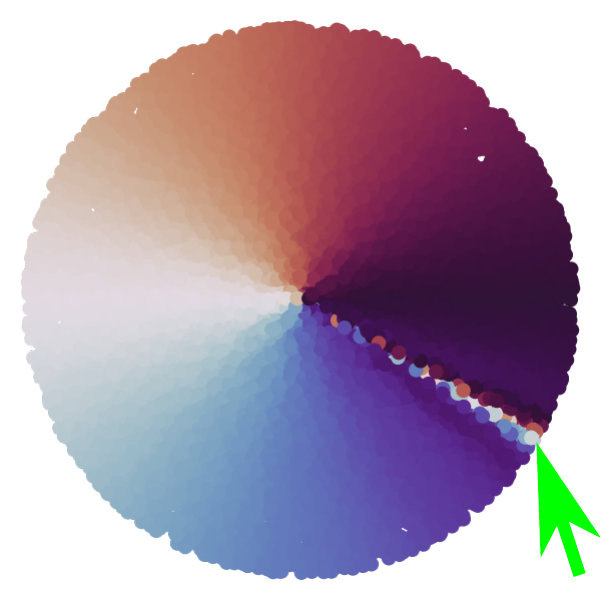}
			& \includegraphics[align=c,width=\myfigwidth]{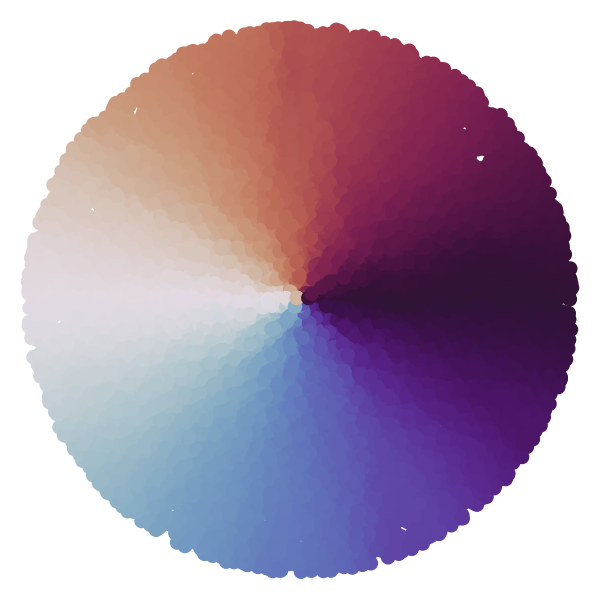}
			& \includegraphics[align=c,width=\myfigwidth]{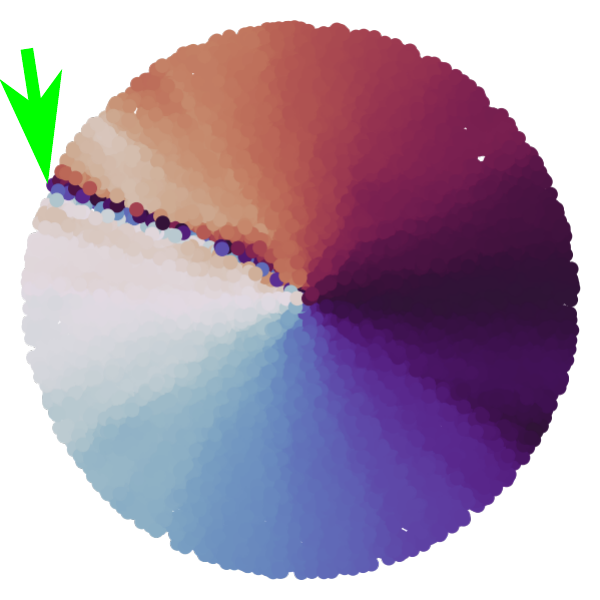}
			& \includegraphics[align=c,width=\myfigwidth]{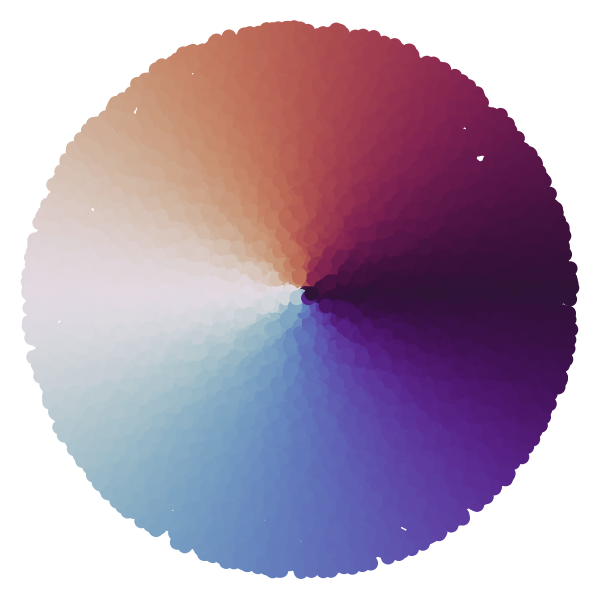}
			&
			\includegraphics[align=c,width=\myfigwidth]{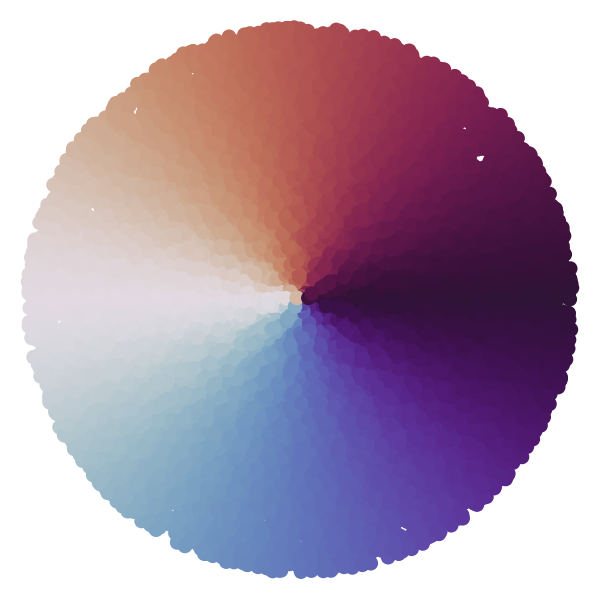}\\		
			\cmidrule[1.5pt]{1-6}	
		\end{tabular}
	}
	{
		\scriptsize
		(a): 1D representation is plotted on an Archimedean spiral for visualization purposes.
		(d): $o$ represents a 2D offset.\\
		$(\dagger)$: Inputs (RGB values of constant lightness) are represented as points on the hue-saturation chromatic disk, with a color corresponding to the predicted hue.
	}
	
	\caption{\label{fig:hue_regression}\textbf{Pre-images connectivity/convexity helps generalization.}
		We train a network to regress color hues from  RGB values. It maps the output of a MLP backbone to the hue circle $Y$ using a function $f$.
		Colors displayed correspond to predicted hues.
		Training can lead the backbone to produce distant representations for training samples sharing a similar hue $y^*$ (green stars, 4$^{th}$ row). If these representations are in disconnected regions of the pre-image $f^{-1}(y^{*})$, the network will not be able to properly interpolate between them (a,b, green marks, last two rows).
		Pre-images connectivity guarantees that there exists a continuous backbone function able to interpolate between such training samples (c), but it can be hard to learn in practice (d). Pre-images convexity (e) provides stronger guarantees and tends to help generalization.}
\vspace{-1em}
\end{figure*}

\noindent\textbf{Pre-images connectivity/convexity: pre-image $f^{-1}(y)$ of any element $y \in Y$ should be connected, or even better convex, to help generalization.}
Indeed, consider the case of a continuous backbone network $h:a \in A \rightarrow x \in X$ producing an intermediate representation $x$ from an input $a$,
and a machine learning task consisting in regressing $y=f(h(a)) \in Y$ on the target manifold close to a target $y^* \in Y$.
We illustrate this with a toy example, consisting in a multi-layer perceptron $h$ taking as input a RGB value $a \in ]0,1[^3$ and regressing an intermediary representation $x = h(a)$ mapped to the hue circle $\mathbb{S}^{1}$. We consider various mapping functions from either a 1D or 2D space depicted in figure~\ref{fig:hue_regression} top row.
Training will hopefully bring the training set representation $x$ close to the pre-image set $f^{-1}(y^*)$ -- a large-enough network being able to fit the training set given a proper training loss function.
However, different inputs $a_0, a_1$ corresponding roughly to the same target output $y^*$ might produce intermediate representations $x_0, x_1$ distant from each other (fig.~\ref{fig:hue_regression}, green stars 4$^{\text{st}}$ row). 
Generalizing requires the model to be able to properly interpolate between training samples.
This is however impossible if $x_0$ and $x_1$ are in disconnected regions of $f^{-1}(y^*)$, as some test representations interpolated by the backbone will necessarily not be mapped to $y^*$ (fig.~\ref{fig:hue_regression}a,b, green marks, last two rows).
The use of a mapping satisfying the pre-images connectivity constraint prevents such situation by guaranteeing that there always exists a backbone function able to properly interpolate between training representations (fig.~\ref{fig:hue_regression}c,e).
Learning such function may however be hard in practice (fig.~\ref{fig:hue_regression}d), and pre-images convexity constraint (fig.~\ref{fig:hue_regression}e) tends to help generalization by guaranteeing that linear interpolation between training representations $x_0, x_1$ will produce the same prediction $y^*$.
Moreover, we show in the supplementary material how the Nash embedding theorem guarantees the existence of a mapping satisfying pre-images connectivity.

\subsection{Softmax example}
Multiple classical deep learning operators are related to this notion of differentiable mapping, and as an example we show how softmax used in classification can be thought of as a function mapping elements of $X=\mathbb{R}^n$ to a probability distribution in $Y=\lbrace (p_i)_{i=1 \ldots n} \in ]0, 1[^n \vert \sum_i p_i = 1 \rbrace$.
$Y$ is a connected $(n-1)$-dimensional differentiable manifold on which softmax is surjective.
Softmax $i$-th component $p_i$ for a vector $(x_1, \ldots, x_n)$ is defined by $p_i = \exp(x_i)/{\sum_j \exp(x_j)}$.
It is differentiable with partial derivatives
$
{\partial p_i}/{\partial x_j} = 
[ p_i(1 - p_j) \text{ if } i=j, 
-p_i p_j  \text{ else} ].
$
The pre-image of any $(p_i)_{i=1 \ldots n} \in Y$ is the convex line
$\softmax^{-1}((p_i)_{i=1 \ldots n}) = \left\lbrace \log(p_i)_{i=1 \ldots n} + c \vert c \in \mathbb{R} \right\rbrace.$
Therefore, for an arbitrary vector $\bm{x} = (x_i)_{i=1 \ldots n}$, one can choose a given $c \in \mathbb{R}$ such as $\log(\softmax(\bm{x})) + c = \bm{x}$, applying $\log$ element-wise.
The differentiable function $g:\bm{p} \in Y \rightarrow \log(\bm{p}) + c$ can be seen as an inverse of the restriction of $f$ to $g(Y)$, forming  a diffeomorphism between $Y$ and $g(Y)$ and therefore guaranteeing the Jacobian of $f$ at $\bm{x}$ to be of same rank as the dimension of $Y$.

\section{\label{sec:deep_representations}Differentiable mappings on SO(3)}

Based on the theoretical considerations developed in section~\ref{sec:mapping_manifold}, we review differentiable functions used in the literature to map arbitrary Euclidean vectors to the 3D rotation space.
Table \ref{tab:constraints_satisfaction} provides a short summary of how these different mappings satisfy the properties listed in section~\ref{sec:mapping_manifold}, and we provide proofs in the supplementary material.
\begin{table}
\caption{\label{tab:constraints_satisfaction}Satisfaction of properties of section~\ref{sec:mapping_manifold} by $SO(3)$ mappings.}
\centering
{
	\footnotesize
	\setlength{\tabcolsep}{2pt}
	\begin{tabular}{|c|cccc|}
\cline{2-5}
\multicolumn{1}{c|}{}		& \makecell{Domain}     & \makecell{Surjective/ \\ differentiable}  & \makecell{Full rank \\ Jacobian}   & \makecell{Connected/convex \\ pre-images} \\
		\hline
		Euler angles        & $\mathbb{R}^3$  & \cmark{} & \xmark{} & \xmark{} \\
		Rotation vector     & $\mathbb{R}^3$  & \cmark{} & \xmark{} & \xmark{} \\
		Quaternion          & $\mathbb{R}^4\setminus\{\bm{0}\}$  & \cmark{} & \cmark{} & \xmark{} \\
		6D                  & $\mathcal{M}_{3,2}(\mathbb{R})^\dagger$  & \cmark{} & \cmark{} & \cmark{} \\
		Procrustes          & $\mathcal{M}_{3,3}(\mathbb{R})^\dagger$  & \cmark{} & \cmark{} & \cmark{} \\
		Symmetric matrix 	& ${\mathbb{R}^{10}}^\dagger$  & \cmark{} & \cmark{} & \cmark{} \\
		\hline
	\end{tabular}\\
	{
	\scriptsize
	$(\dagger)$: defined on all but a null-measure subset of the domain.
	}
}
\vspace{-1em}
\end{table}

\paragraph{Euler angles}
Euler~\cite{euler_formulae_1776} showed that the orientation of any rigid body could be expressed by 3 angles describing a succession of 3 rotations around elementary axes (the choice of which being a matter of convention).
As an example, one can consider a succession of rotations around respectively the $x-y-z$ axes and define a mapping
$(\alpha, \beta, \gamma) \in \mathbb{R}^3 \rightarrow \bm{R}_x(\alpha) \bm{R}_y(\beta) \bm{R}_z(\gamma) \in SO(3).$
While being surjective, it does not satisfy pre-images connectivity in general due to the existence of multiple discrete pre-images for a given rotation, and its Jacobian suffers from rank deficiency for some rotations (phenomenon referred to as \emph{gimbal lock}).
Zhou~\etal{}~\cite{zhou_unsupervised_2017} \eg predict a 3D translation vector and a rotation parameterized by 3 Euler angles, latter cast into a rigid transformation matrix as part of their unsupervised training strategy.

\paragraph{Rotation vector} Rotation vector space can be seen as an unrolling of the rotation space on a tangent plane of the identity rotation.
Any arbitrary 3D vector can be mapped to the rotation space through the exponential map (\eg using Rodrigues' formula), which consists in a surjective function over the rotation space.
However, similarly to Euler angles it suffers from multiple discrete pre-images, and rank deficiency for input rotation vectors of angle $2\pi k, k \in \mathbb{N}^*$.
It is used \eg in \cite{rong_delving_2019} to regress parameters of a human model.
Note however that a restriction from an open ball of radius $\alpha < \pi$ to the set of rotations of angle strictly smaller than $\alpha$ satisfies all properties of section \ref{sec:mapping_manifold}, and that such mapping is therefore suited to regress rotations of limited angles.

\paragraph{Non zero quaternion}
The rotation group $SO(3)$ is diffeomorphic to $\mathbb{RP}^3$~\cite{hall_2015} and therefore one can define a mapping from $X=\mathbb{R}^4\setminus \lbrace \bm{0} \rbrace$ to $SO(3)$ by considering each element $\bm{x} \in X$ as a representation of a rotation parameterized by a unit norm quaternion $\bm{q}=\bm{x}/\|\bm{x}\|$.
Such mapping satisfies all but pre-images connectivity constraint as the pre-image of any rotation parameterized by a non zero quaternion $\bm{q}$ consists in 
$\lbrace \alpha \bm{q} \vert \alpha \in \mathbb{R}^* \rbrace$
which is not connected.
This mapping is used \eg in PoseNet~\cite{kendall_posenet_2015}, and Liao~\etal{}~\cite{liao_spherical_2019} suggest a variant of this mapping to stabilize training.

\paragraph{Procrustes} 
An arbitrary $3 \times 3$ matrix $\bm{M}$ can be projected to the closest rotation matrix considering Frobenius norm by solving the special orthogonal Procrustes problem
$\procrustes(\bm{M}) = \arg\min_{\bm{\tilde{R}} \in SO(3)} \| \bm{\tilde{R}} - \bm{M} \|_F^2.
$
Solutions to this minimization problem 
 can be 
expressed $\bm{U} \bm{S} \bm{V}^\top$~\cite{schonemann_generalized_1966, umeyama_least-squares_1991},
where $\bm{U} \bm{D} \bm{V}^\top$ is a singular value decomposition of $\bm{M}$
such that $\bm{U}, \bm{V} \in O(d)$, $\bm{D} = \diag(\alpha_1,\alpha_2, \alpha_3)$ is a diagonal matrix satisfying
$\alpha_1 \geq \alpha_2 \geq \alpha_3 \geq 0$
and $\bm{S}$ is defined by $\diag(1,1, \det(\bm{U}) \det(\bm{V}))$.
The solution is actually unique if $\det(\bm{M}) > 0$ or $\alpha_2 \neq \alpha_3$, and \emph{Procrustes} mapping is therefore well-defined almost everywhere on the set of $3 \times 3$ matrices, but on a set of null measure (usually ignored in practice). We show in the supplementary material that it satisfies all properties presented in section~\ref{sec:mapping_manifold}, notably that the pre-image of any rotation is convex. \emph{Procrustes} mapping is therefore a suitable candidate for reliable learning, and it can be implemented efficiently for batched GPU applications using off-the-shelf SVD \emph{cuSOLVER} routines.
It is used \eg in \cite{omran_neural_2018} for human pose estimation, and some concurrent work~\cite{levinson_analysis_2020} advocates for its use as a universally effective representation.

\paragraph{6D}
Zhou \etal~\cite{zhou_continuity_2019} proposed a similar approach consisting in mapping a $3 \times 2$ matrix $\bm{M}$ (6D in total) to a rotation matrix through Gram-Schmidt orthonormalization.
We show in supplementary material that this process can be thought of as a degenerate case of Procrustes orthonormalization, expressed as a limit
\begin{equation}
\label{eq:gramschmidt_as_procrustes_limit}
\SixD(\bm{M}) = \lim_{\alpha \rightarrow 0^+} \arg\min_{\bm{R} \in SO(3)}  \| \bm{R} \left( \begin{smallmatrix} 1 & 0 \\ 0 & \alpha \\ 0 & 0\end{smallmatrix} \right) - \bm{M}  \|_F^2.
\end{equation}
Solving this optimization problem for a given $\alpha > 0$ is equivalent to solving 
the \emph{Procrustes} one, considering $\bm{M} \left( \begin{smallmatrix} 1 & 0 & 0\\ 0 & \alpha & 0 \end{smallmatrix} \right) \in \mathcal{M}_{3,3}(\mathbb{R})$ as input. Both mappings therefore share in that sense similar properties, except that the \emph{6D} mapping gives importance almost exclusively to the first column of $\bm{M}$ and is only well-defined when $\rank(\bm{M}) = 2$.
Zhou~\etal{}~\cite{zhou_continuity_2019} showed that this 6D representation could actually be compressed into a 5D one through the use of stereographic projection, which led to worse training performances in practice.
We show in supplementary material how this mapping satisfies properties of section~\ref{sec:mapping_manifold}.
The \emph{6D} mapping is used in various recent papers~\cite{ pavlakos_texturepose_2019, kocabas_vibe_2020}, but experiments  of \cite{levinson_analysis_2020} and ours suggest that it is generally outperformed by Procrustes orthonormalization.

\noindent\textbf{Symmetric matrix}
Lastly, Peretroukhin~\etal{}~\cite{peretroukhin_smooth_2020} recently proposed to regress 3D rotations as a list of 10 coefficients of a $4 \times 4$ symmetric matrix $\bm{A}$, and showed that one could map such matrix to a unique rotation (expressed as a unit quaternion) associated with a particular probability distribution, as long as the smallest eigenvalue of $\bm{A}$ has a multiplicity of 1.
We show in the supplementary material how it also satisfies the properties of section~\ref{sec:mapping_manifold}.

\paragraph{\emph{RoMa} library} Writing code handling various rotation representations and mappings can be hard and error-prone because of the multiple corner-cases to consider.
To ease deep learning experiments involving these rotation mappings, we release \emph{RoMa} ({\small{}\url{https://naver.github.io/roma/}}) -- an easy-to-use and efficient PyTorch library for rotation manipulation, and hope it will support future research.

\section{\label{sec:experiments}Experiments}
We report in this section a methodological performance evaluation of $SO(3)$ mappings on a variety of tasks,
namely point cloud alignment (sec.~\ref{subsec:pointcloud}), inverse kinematics (sec.~\ref{subsec:ik}), camera localization (sec.~\ref{subsec:camera_localization}) 
and object pose estimation (sec.~\ref{subsec:object_pose}).
More precisely, we present in sections~\ref{subsec:pointcloud} and \ref{subsec:ik} a reproduction and extension of experiments of \cite{zhou_continuity_2019} including an additional mapping \emph{Procrustes}, and a study of the results variance across multiple trainings.
Benchmarks proposed in sections~\ref{subsec:camera_localization} and ~\ref{subsec:object_pose} consists in novel tasks.

The goal of these experiments is twofold: 1) test the relevance of properties proposed in section~\ref{sec:mapping_manifold} regarding what makes a good mapping, and 2) provide a performance comparison of $SO(3)$ mappings on a variety of tasks.

We did not include Euler angles in our experiments as many different conventions would have to be considered.
Similarly, we ignored the mapping based on a $4 \times 4$ symmetric matrix representation~\cite{peretroukhin_smooth_2020}, as its implementation was found too slow to be used in extensive evaluations.
When the loss function is expressed directly as a distance in $3 \times 3$ matrix space (sections~\ref{subsec:pointcloud} and \ref{subsec:object_pose}), we additionally evaluate variants consisting in regressing an arbitrary $3 \times 3$ matrix at training time using this particular loss, and performing the mapping on SO(3) only at test time (we refer to these variants as \emph{Matrix/Procrustes} and \emph{Matrix/Gram-Schmidt} depending on the mapping used at test time).

\subsection{\label{subsec:pointcloud}Point cloud alignment}
We reproduce and extend the experiment of \cite{zhou_continuity_2019} on point clouds alignment, which while being artificial, provides an interesting benchmark.
Given an input 3D point cloud $C_1=\lbrace \bm{x}_i \rbrace_{i=1 \ldots n}$ of size $n$ (with $\bm{x}_i \in \mathbb{R}^3$), and a copy $C_2=\lbrace r^*(\bm{x}_i) \rbrace_{i=1 \ldots n}$ rotated by a random rotation $r^* \in SO(3)$, we train a neural network to regress this rotation.
The network consists of two Siamese naive PointNets returning a feature vector for each point cloud, which are concatenated and passed through a multi-layer perceptron to produce a feature vector mapped to a rotation $r$ using one of the mappings studied.
Consistently with Zhou~\etal, training is performed by minimizing the square Frobenius distance between rotation matrices $\bm{R}, \bm{R}^* \in \mathcal{M}_{3,3}(\mathbb{R})$ representing these rotations $r$ and $r^*$:
$\mathcal{L}(r, r^*) = \| \bm{R} - \bm{R}^* \|_F^2$.

\begin{figure}
\centering
\noindent\hspace{-1em}\includegraphics[scale=1.0]{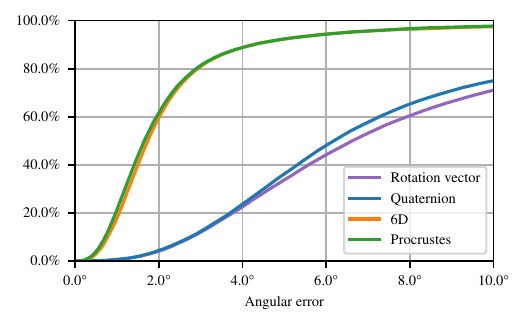}
\vspace{-22pt}
\caption{\label{fig:shapenet_error_distribution}Cumulative angular error distribution on test set for point cloud alignment (best validation models only, for readability).}
\vspace{-1em}
\end{figure}

\paragraph{Protocol}
We follow the protocol of \cite{zhou_continuity_2019}, using the dataset provided by the authors representing airplanes from ShapeNet~\cite{shapenet2015}.
We select however randomly a validation set of 400 point clouds from the original training set, and randomly subsample point clouds to a size of 1024.
We run the experiment with various mappings to the rotation space and report results in table~\ref{shapenet_results} and figure~\ref{fig:shapenet_error_distribution}.
Due to the stochastic nature of training,
we train/test each variant 5 times and aggregate results to mitigate their variance and increase confidence in our findings.
More precisely, we report the error for the model having obtained the smallest validation error (being checked after each training epoch), as well as the final error obtained at the end of training, averaged over 5 runs.

\begin{table}
\centering
\caption{\label{shapenet_results}Mean angular error on test set  for point cloud alignment, aggregated across 5 independent trainings  (lower is better).}
{
\footnotesize
{\scriptsize Subscript and exponent represent the min.\ and max.\ deviations from the average.}
\renewcommand{\arraystretch}{1.3}
\setlength{\tabcolsep}{3pt}
\begin{tabular}{|c|cc|}
\hline
Method           & Best validation & Average final error (\textdegree{}) \\
\hline
Rotation vector           & $12.74\degree{}$            & $14.27_{-0.84}^{+1.32}$ \\
Quaternion                & $11.66\degree{}$                 & $12.99_{-1.13}^{+1.49}$ \\
6D                        & $3.01\degree{}$                  & $3.47_{-0.45}^{+0.45}$  \\
\textbf{Procrustes}       & $\bm{2.90\degree}$             & $\bm{3.46}_{-0.35}^{+0.59}$  \\
\hline
Matrix/Gram-Schmidt     & $3.66\degree$                  & $4.07_{-0.10}^{+0.21}$ \\
\textbf{Matrix/Procrustes}       & $\bm{3.33\degree}$             & $\bm{3.59}_{-0.09}^{+0.08}$ \\
\hline
\end{tabular}
}
\vspace{-1em}
\end{table}

\paragraph{Results}
In accordance with the results reported in~\cite{zhou_continuity_2019}, \emph{rotation vector} and \emph{quaternion} mappings achieve significantly worse performances than the other mappings.
Procrustes orthonormalization leads to better results than Gram-Schmidt one, both during training (\emph{Procrustes} \vs \emph{6D}), and when used only at test time (\emph{Matrix/Procrustes} \vs \emph{Matrix/Gram-Schmidt}).
Difference of average final error between \emph{Procrustes} and \emph{6D} is too small however to draw clear conclusions from this metric alone, given the deviations observed across the different runs.

Interestingly, we achieve better performance when performing orthonormalization both at training and test time (\emph{6D} and \emph{Procrustes}) compared to regressing a raw matrix and performing orthonormalization only for validation/test (\emph{Matrix/Gram-Schmidt} and \emph{Matrix/Procrustes}).
However, we observe a smaller final error variance when training without orthonormalization, with a maximum deviation of 0.18\textdegree{} for \emph{Matrix/Procrustes} \vs 0.91\textdegree{} for \emph{Procrustes}, and 0.31\textdegree{} for \emph{Matrix/Gram-Schmidt} \vs 0.89\textdegree{} for \emph{6D}.
We explain this by the fact that learning is more constrained in the direct regression scenario.

\subsection{\label{subsec:ik}Inverse kinematics}

We reproduce and extend another experimental setup of~\cite{zhou_continuity_2019}, more representative of a real use case, and where rotation regression is trained in a self-supervised manner.

The problem consists in learning an inverse kinematic model of a human skeleton from a set of recorded motions.
The forward kinematic function $f$ mapping the $n=57$ rotations $(r_i)_{i=1 \ldots n} \in SO(3)^n$ associated to each joint 
to their 3D locations $(\bm{x}_i)_{i=1 \ldots n} \in \mathbb{R}^{3 \times n}$ is assumed to be known.
We train a multi-layer perceptron $g:\mathbb{R}^{3 \times n} \rightarrow SO(3)^n$ to regress the inverse mapping.
During training, given a target list of 3D joints locations $(\bm{x}^*_i)_{i=1 \ldots n}$, we use an auto-encoding approach regressing a list of joints locations
$(\bm{x}_i)_{i=1 \ldots n} = f(g((\bm{x}^*_i)_{i=1 \ldots n})$,
and train the network to minimize the square residual
$\sum_{i=1}^n \alpha_i \| \bm{x}_i - \bm{x}^*_i \|^2$.

\paragraph{Protocol}
We follow closely the protocol of~\cite{zhou_continuity_2019} and use data provided by the authors from the CMU MoCap Database~\cite{cmu_mocap}. 
We select randomly 73 training sequences as validation set, and train each variant with 5 different random weight initializations to assess the confidence of our observations.

\begin{figure}
\centering
\noindent\hspace{-1em}\includegraphics[scale=1.0]{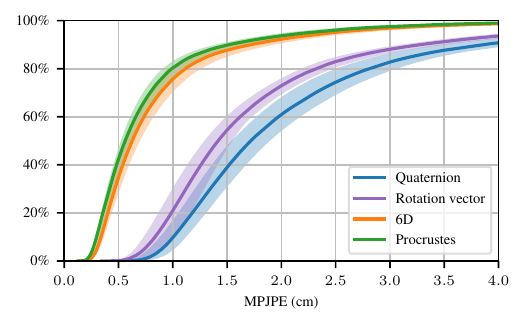}
\vspace{-22pt}
\caption{\label{fig:ik_plot}Cumulative mean per joint error distribution on the inverse kinematics test set, aggregated across 5 independent trainings (Solid line: average distribution, shaded areas: maximal deviation across runs).}
\vspace{-1em}
\end{figure}

\paragraph{Results}
We report results in table~\ref{tab:ik} and figure~\ref{fig:ik_plot}. The $\emph{Procrustes}$ and $\emph{6D}$ mappings greatly outperform the others in this experiment, with $\emph{Procrustes}$ showing significantly better performances than $\emph{6D}$.

\begin{table}
\centering
\caption{\label{tab:ik}Mean per joint position error on the inverse kinematics test set, aggregated across 5 independent trainings (lower is better).}
{
\footnotesize
{\scriptsize Subscript and exponent represent the min.\ and max.\ deviations from the average.}
\renewcommand{\arraystretch}{1.2}
\begin{tabular}{|c|cc|}
\hline
Method              & Best validation & Average final error (cm)\\
\hline
Quaternion          & 1.948cm & $2.208_{-0.241}^{+0.208}$ \\ 
Rotation vector     & 1.695cm & $1.840_{-0.144}^{0.102}$ \\
6D                  & 0.796cm & $0.885_{-0.079}^{+0.103}$ \\
\textbf{Procrustes} & \textbf{0.721cm} & $\bm{0.795}_{-0.061}^{+0.088}$ \\
\hline
\end{tabular}
}

\vspace{-1em}
\end{table}

\subsection{\label{subsec:camera_localization}Camera localization}

We study the impact of mapping function on the problem of absolute 6D camera localization, and base our experiments on PoseNet~\cite{kendall_posenet_2015},  a pioneer work of image-based direct camera pose regression.
%
PoseNet processes an input RGB image using a CNN to regress the camera pose from which the image was taken, after a prior supervised training.
In its original version, camera pose is regressed as a pair $(\bm{t}, \bm{q})$ composed of a translation vector $\bm{t} \in \mathbb{R}^3$ and a unit quaternion $\bm{q} \in \mathbb{R}^4$ obtained by normalizing an arbitrary 4D output of a neural network.
Given a ground truth pose $(\bm{t}^*, \bm{q}^*)$, original training aims to minimize the loss function
\begin{equation}
\label{eq:posenet_original_loss}
\mathcal{L}_{original} = \alpha \|\bm{t} - \bm{t}^* \|^2 + \beta \gamma \| \bm{q} - \bm{q}^* \|^2,
\end{equation}
with $\alpha = 1/3$, $\gamma = 1/4$ and where $\beta$ is a scaling factor weighting the importance of rotation \vs translation.

\paragraph{Protocol}
We replace the last fully connected layer of the network to produce a feature vector of adequate dimension to apply each evaluated mapping and regress a rotation.
Rotation part of the original loss function~\eqref{eq:posenet_original_loss} is actually biased as it is not a valid similarity measure on $SO(3)$.
Indeed, a 3D rotation can be represented by two antipodal unit quaternions $\bm{q}$ and $-\bm{q}$, therefore one should instead consider
\begin{equation}
\label{eq:posenet_quaternion_loss}
\mathcal{L}_{quaternion} = \alpha \|\bm{t} - \bm{t}^* \|^2 + \beta \gamma \min \| \bm{q} \pm \bm{q}^* \|^2
\end{equation}
in order to avoid border effects of the representation.

To investigate the impact of the training loss function on the results, we also experiment to replace it by one based on Frobenius distance between rotation matrices $\bm{R}, \bm{R}^*$
\begin{equation}
\label{eq:posenet_frobenius_loss}
\mathcal{L}_{Frobenius} = \alpha \|\bm{t} - \bm{t}^* \|^2 + \beta \delta \| \bm{R} - \bm{R}^* \|_F^2.
\end{equation}
We choose $\delta = \gamma/8$ to weight the losses similarly (see supp. mat. for details).
We evaluate variants of PoseNet on \emph{Cambridge Landmarks} datasets, and report median camera position and orientation errors.
We use the PoseNet implementation of Walch~\etal~\cite{walch_image-based_2017} with its recommended hyperparameters settings.
We observed negligible variance in the results across different training, because of the use of pre-trained weights and deterministic seeding,
and we therefore report a single result for each mapping.
 
\paragraph{Results}
The results obtained are summarized in table~\ref{tab:posenet_results}.
\begin{table*}
\centering
\caption{\label{tab:posenet_results}Median camera localization error on \emph{Cambridge Landmarks} datasets.}
{
	\small
	\setlength{\tabcolsep}{1pt}
	\input{merged_posenet_table}

}
\end{table*}
The baseline \emph{quaternion} mapping trained with the original loss~\eqref{eq:posenet_original_loss} performs well compared to the variants trained with the quaternion loss~\eqref{eq:posenet_quaternion_loss}.
It is not totally surprising in the sense that hyperparameters have been tuned for this particular setting, introducing a bias in its favor.
It notably performs better on average than the \emph{quaternion} mapping trained using loss~\eqref{eq:posenet_quaternion_loss}, despite the fact the former is ill-defined on $SO(3)$ as described previously.
We conjecture that it might be due to the combination of two factors. 
First, loss~\eqref{eq:posenet_quaternion_loss} might be confusing during training and lead to poorer generalization than loss~\eqref{eq:posenet_original_loss} when using a quaternion representation because 4D outputs of the backbone network corresponding to nearby orientations may be pushed towards different and opposite regions with this loss (this relates to the fact that the \emph{quaternion} mapping does not satisfy the pre-images connectivity property).
Second, ground truth unit quaternions representations $\bm{q}^*$ are oriented consistently in the dataset. For each dataset more than 99\% of ground truth quaternions are indeed oriented in the same half-space, mostly far from its border, which mitigates the potential border effects introduced by loss~\eqref{eq:posenet_original_loss}. 

Results obtained using the \emph{Procrustes} representation are globally better than those using the \emph{6D} or \emph{quaternion} representations for both losses~\eqref{eq:posenet_quaternion_loss} and~\eqref{eq:posenet_frobenius_loss}, and are on average better than the baseline.

The \emph{rotation vector} representation performed surprisingly well in this experiment despite the fact that this mapping does not satisfy the criteria of section~\ref{sec:mapping_manifold}, even achieving the best average results for loss~\eqref{eq:posenet_frobenius_loss}. We conjecture that this success might be due to a combination of a `lucky' weight initialization and to the fact that rotations to regress in the datasets are typically less than 90\textdegree{} away from the identity, enabling the use of a portion of the rotation vector space in which properties of section~\ref{sec:mapping_manifold} are roughly met (a similar hypothesis was studied in~\cite{parascandolo_taming_2016} for non-monotonic activation functions).
We tested this hypothesis by applying to the regressed rotation a 180\textdegree{} rotation around the $x$ camera axis before the evaluation of the loss, in such a way that poses to regress are no longer close to the identity. Training completely failed to converge to a good solution as reported in table~\ref{tab:posenet_results} for loss~\eqref{eq:posenet_quaternion_loss}, which supports our hypothesis.
It did however succeed for loss~\eqref{eq:posenet_frobenius_loss}, reaching performances of the order of those of the non-rotated version.

\subsection{\label{subsec:object_pose}Object pose estimation}

Lastly, we investigate the impact of the mapping function on object pose estimation from an RGB image.

\paragraph{Protocol}
We try to predict the 3D orientation of a rigid object from an image crop extracted from LINEMOD dataset~\cite{hinterstoisser_multimodal_2011}.
Following a common practice~\cite{rad_bb8_2017}, we use synthetic data for training, rendering 200,000 crops for each object with randomized pose, illumination and background.
For the sake of simplicity, we do not use any real data, physically-based renderings, or data-augmentation during training and cannot therefore pretend reaching the same level of performances than state-of-the-art methods~\cite{hodan_bop_2020}.
We use for each object a pre-trained ResNet-50 backbone~\cite{he_deep_2016},
and replace the last fully connected layer by one whose output dimension is suitable for the mapping considered.
We freeze the first layers of the backbone (up to the second one) to speed up the training and following insights from~\cite{hinterstoisser_pre-trained_2018} regarding generalization performances.
Given a ground truth object orientation expressed as a rotation $r^* \in SO(3)$, we train the model to regress a rotation $r \in SO(3)$ minimizing the mean square position error of the object's surface points
\begin{equation}
\label{eq:object_loss}
\mathcal{L}(r, r^*)=\cfrac{1}{d^2 S} \int_{\mathcal{S}}{ \|r(\bm{x}) - r^*(\bm{x}) \|^2 ds},
\end{equation}
where $\mathcal{S}$ represents the surface of the object expressed in a coordinate system centered on its centroid, $S$ its area and where $d$ is a normalizing factor corresponding to the diameter of the object.
We showed in~\cite{bregier_defining_2018} that this loss function admits a  closed-form solution 
$\| (\bm{R} - \bm{R}^*) \bm{\Lambda} \|_F^2$,
where $\bm{\Lambda}$ is a symmetric positive matrix depending on the object.
We train the network for 30 epochs using SGD and perform hyperparameters search on the validation set (see supp. mat. for details).
At test time, we report the mean RMS error regarding the position of the surface points.

\paragraph{Results}
The results are summarized in table~\ref{tab:object_results}. The \emph{Procrustes} and \emph{6D} mappings achieve better results than the \emph{quaternion} and  \emph{rotation vector} ones, and \emph{Procrustes} outperforms \emph{6D} for most objects.
Training to regress an arbitrary $3 \times 3$ matrix without special orthonormality constraints leads to worse performances at test time, but Procrustes orthonormalization still outperforms Gram-Schmidt in this scenario.

\begin{table}
\centering
\caption{\label{tab:object_results} Regression of object orientation from an image. Mean RMS error normalized by the object diameter (lower is better).}
{
\footnotesize
\setlength{\tabcolsep}{2pt}
\input{object_table.tex}
}
\vspace{-1em}
\end{table}

\section{\label{sec:discussion}Discussion}

In this section, we try to derive some general insights from the results of our experiments.

\noindent{}\textbf{Theory validation}
We globally observed better results with the \emph{Procrustes} and \emph{6D} mappings than with the ones based on \emph{quaternion} and \emph{rotation vectors}.
This is consistent with the theory developed in section~\ref{sec:mapping_manifold} regarding what properties should satisfy a mapping to enable good training and generalization.
In the camera localization experiment, the drop of performance observed with the \emph{quaternion} mapping when switching from the original PoseNet loss~\eqref{eq:posenet_original_loss} to a well-defined loss on the rotation space~\eqref{eq:posenet_quaternion_loss} supports the idea that the non-connectivity of pre-images is a problem for learning.
It is somehow also supported by the failure to converge to a low error solution observed with the \emph{Rot-vec 180\textdegree{}} representation.
We tested our theory on several manifolds -- $\mathbb{S}^1$, $SO(3)$, $SO(3)^{n}$, and $SE(3)$ (resp. in fig.~\ref{fig:hue_regression}, sections~\ref{subsec:pointcloud} and~\ref{subsec:object_pose}, \ref{subsec:ik}, \ref{subsec:camera_localization}) -- and we also report supporting results on a torus as supplementary material.

\begin{figure}
\centering
\vspace{-5pt}
\noindent\hspace{-1em}\includegraphics[scale=1.0]{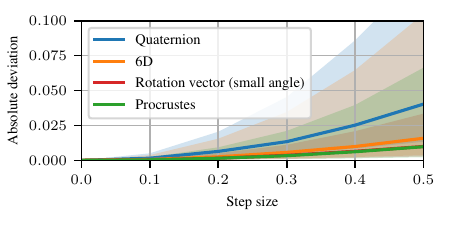}

{
\vspace{-10pt}
\scriptsize
\centering{}
Curve corresponding to \emph{Rotation vector} is masked by \emph{Procrustes}.\\
}
\caption{\label{fig:linearity}Deviation from linearity of different mappings on SO(3) (solid curves: median, shaded areas: 25\%-75\% percentiles).}
\end{figure}

\noindent{}\textbf{Mapping `linearity'}
We also observed that regressing a $3 \times 3$ matrix and performing \emph{Procrustes} orthonormalization at training time globally performed better than using the \emph{6D} representation.
Work concurrent to ours~\cite{levinson_analysis_2020} led to similar conclusions and provides two arguments. The first consists in the smaller sensitivity of Procrustes orthonormalization to Gaussian noise compared to Gram-Schmidt, but their results only hold for local perturbations applied to a rotation matrix. The second is that Procrustes orthonormalization is left- and right-invariant to rotations, whereas Gram-Schmidt is only left-invariant (we discussed in section~\ref{sec:deep_representations} how \emph{6D} gives more importance to the first input column vector).
We propose here as another explanation that the \emph{Procrustes} mapping is more `linear' than $\emph{6D}$ and \emph{quaternion} for typical input ranges, and we support this  claim by numerical experiments.
We draw for each mapping $\bm{R}: \bm{x} \in \mathbb{R}^n \rightarrow \bm{R}(\bm{x}) \in SO(3)$ some random input vector
$\bm{x} \sim \mathcal{N}(0,1)^n$, as well as some unit 3D vectors $\bm{v}_1, \bm{v}_2$ of uniform random direction.
Because of the way \emph{Procrustes}, \emph{6D} and \emph{quaternion} mappings are defined, $\bm{R}(\bm{x})$ is uniformly distributed over $SO(3)$ and no bias is introduced by this approach. 
We define a loss function $\mathcal{L}(\bm{x}) = \bm{v}_1^T \bm{R}(\bm{x}) \bm{v}_2$, estimate its gradient $\nabla \mathcal{L}_{\bm{x}}$ and compute for a given step size $\epsilon > 0$ the absolute deviation from the linear case:
$| \mathcal{L}(\bm{x} - \epsilon \nabla \mathcal{L}_{\bm{x}}) - (\mathcal{L}(\bm{x}) - \epsilon \|\nabla \mathcal{L}_{\bm{x}}\|^2) |$.
Results for various step sizes are plotted in fig.~\ref{fig:linearity} and show the better linearity of \emph{Procrustes} compared to \emph{quaternion} and \emph{6D}.

\noindent{}\textbf{Direct/surrogate objective} We also observed that using a differentiable mapping $\bm{x} \rightarrow f(\bm{x})$ at training time and defining a loss on the target manifold performed better in our experiments than training to regress a representation $\bm{x}$, and mapping it to the manifold at test time (\eg \emph{Procrustes} \vs \emph{Matrix/Procrustes}).
This seems a natural finding as we optimize in the former case an objective function corresponding to our actual goal defined on the manifold, whereas the latter case considers only a surrogate objective in the representation space.

\noindent{}\textbf{Small rotation vector}
Finally, our experiments with PoseNet showed that a \emph{rotation vector} mapping could perform well in some circumstances, and notably when considering rotations of limited angle.
Such mapping actually satisfies the criteria of section~\ref{sec:mapping_manifold} when restricted from a unit ball of radius $\alpha \in ]0, \pi[$ to the corresponding set of rotations, and the plot in figure~\ref{fig:linearity} indeed suggests a linearity similar between \emph{Procrustes} and the mapping to rotations of angle strictly smaller than $\pi/2$ 
{
\small
\begin{equation}
\label{eq:rotation-vector-small-angle}
\bm{x} \in \mathbb{R}^3 \rightarrow \exp ( \cfrac{\pi}{2} \tanh(\|\bm{x}\|) \cfrac{\bm{x}}{\|\bm{x}\|} )
\end{equation}
}
for normal inputs\footnote{Normal inputs enable a reasonable coverage of the target output space.}.

\section{Conclusion}
In this paper, we study the problem of deep regression on a manifold through the use of a mapping from the Euclidean output space of a neural network to this manifold.
We establish a list of properties that such mapping should satisfy to allow a proper gradient-based training, and highlight in particular the importance of pre-images connexity/convexity.
We review the specific case of the 3D rotation manifold, considering existing mappings both from a theoretical and experimental standpoint, and conjecture that linearity of the mapping might be an important additional aspect to achieve good performances.
We show that a mapping based on special Procrustes orthonormalization performs best among the mappings considered when regressing arbitrary rotations,
but that rotation-vector representations may be as suitable when the output can be constrained to limited rotation angles.

{\small
	\bibliographystyle{ieee_fullname}
	\bibliography{biblio}
}

\end{document}


\title{Deep Regression on Manifolds: A 3D Rotation Case Study \\ Supplementary material}

\author{Romain Br\'egier\\
	NAVER LABS Europe\\
	{\tt\small romain.bregier@naverlabs.com}
}
	
\maketitle
	
This document provides some supplementary material to the paper \emph{Deep Regression on Manifolds: A 3D Rotation Case Study}.
Section~\ref{sec:torus} describes an additional regression experiment performed on a torus.
Section~\ref{sec:experimental_details} provides some technical details regarding the experiments described in the paper, and section~\ref{sec:proof} provides proofs regarding mathematical statements contained in the paper.
Lastly, we provide in the file \emph{hue\_regression\_example.py} the Python code used to generate figures of our hue regression example.

\section{\label{sec:torus}Regression on a torus}
\begin{figure}
	\footnotesize
	
	(a) \includegraphics[align=c,height=1.5cm]{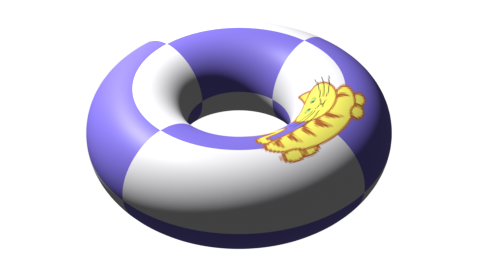}
	\includegraphics[align=c,height=1.5cm]{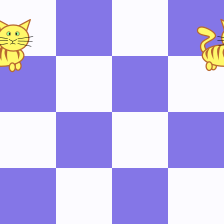}\\
	
	(b)
	\begin{tabular}{c|cc}
		\includegraphics[height=1.3cm]{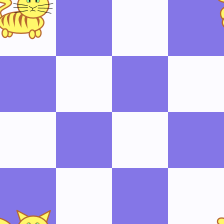}
		& \includegraphics[height=1.3cm]{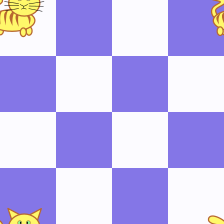}
		& \includegraphics[height=1.3cm]{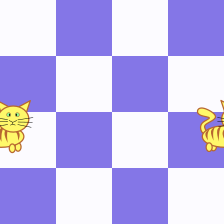} \\
		Input & \emph{atan2} prediction & \emph{linear} prediction
	\end{tabular}\\
	\caption{\label{fig:torus_exp}(a) Regression of the position of a character on a torus (left) given an input RGB map (right). (b) Qualitative results.}
\end{figure}

In this section, we present a toy experiment illustrated in figure~\ref{fig:torus_exp}. It consists in regressing the position of a character on a \emph {PacMan}-like torus $\mathbb{S}^1 \times \mathbb{S}^1$ from an input RGB map.
We represent a point on the torus by a 2D vector associated with the following distance:
\begin{equation}
\forall \bm{y}, \bm{y}^* \in \mathbb{R}^2,\text{ } d(\bm{y}, \bm{y}^*) \triangleq \|(\bm{y} - \bm{y}^*) - [\bm{y} - \bm{y}^*]\|, 
\end{equation}
where $[ \cdot ]$ denotes the element-wise rounding operator.

We use a ResNet18 backbone to extract a global representation $\bm{x} \in \mathbb{R}^n$ of the input image (replacing the last FC layer of the backbone by one of appropriate dimension), and map it to the torus using one of the following mappings:
{
\footnotesize{}
\begin{equation}
\left\lbrace
\begin{aligned}
&\text{\emph{identity}}: \bm{x} \in \mathbb{R}^2 \rightarrow \bm{x} \\
&\begin{aligned}
&\text{\emph{atan2}}: &(x_1, \dots, x_4)  \in \mathbb{R}^4 \setminus \left( \lbrace(0,0)\rbrace \times \mathbb{R}^2 \cup \mathbb{R}^2 \times \lbrace(0,0)\rbrace \right) \\
&& \rightarrow  \left( \cfrac{\atan2(x_2, x_1)}{2\pi}, \cfrac{\atan2(x_3, x_2)}{2\pi}\right).
\end{aligned}
\end{aligned}
\right.
\end{equation}
}

Both mappings are surjective and differentiable with a full rank Jacobian but only \emph{atan2} satisfies the properties of pre-images connectivity/convexity.
Indeed, for any $\bm{x}$ in the domain of \emph{atan2}, the pre-image $\text{\emph{atan2}}^{-1}(\text{\emph{atan2}}(\bm{x}))$ is the convex set $\lbrace \alpha \bm{x} \vert \alpha > 0 \rbrace$, whereas for any $\bm{x} \in \mathbb{R}^2$ the pre-image $\text{\emph{identity}}^{-1}(\text{\emph{identity}}(\bm{x}))$ is the disconnected set $\lbrace \bm{x} + \alpha (0,1) + \beta (1, 0) \vert \alpha, \beta \in \mathbb{Z} \rbrace$.

We use the square distance between predicted and target position as training loss.
We train the network for 25 epochs using Adam with a learning rate of $10^{-4}$ on a training set of 1000 random samples. We test it on a dataset of similar size. 

Quantitative and qualitative results in table~\ref{tab:torus_exp} and figure~\ref{fig:torus_exp} show that \emph{atan2} performs significantly better than \emph{identity} both on the training set and the test set, which supports our theory. \emph{Identity} notably suffers from severe overfitting, with a mean position error more than 4 times higher on the test set than on the training set.

\begin{table}
	\caption{\label{tab:torus_exp}Quantitative results for regression on a torus..}
	\centering
	{
		\footnotesize
		\begin{tabular}{ccccc}
			\toprule{}
			Mapping & \multicolumn{2}{c}{Mean loss ($\times 10^{-4}$)} & \multicolumn{2}{c}{Mean error ($\times 10^{-2}$)} \\
			& train & test & train & test \\
			\midrule{}
			\emph{identity} & 32 & 607 & 4.9 & 21 \\
			\emph{atan2} & 1.6 & 5.0 & 1.1 & 1.9 \\
			\bottomrule{}
		\end{tabular}
	}
\end{table}

\section{\label{sec:experimental_details}Experimental details}

\begin{figure}
	\centering
	\includegraphics[height=1.2in]{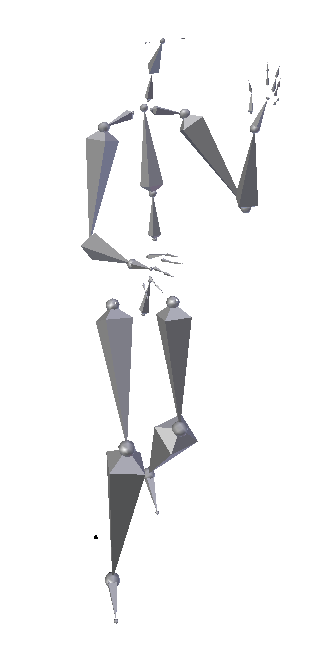}
	\hfill
	\includegraphics[height=1.2in]{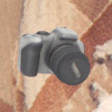}
	\hfill	
	\includegraphics[height=1.2in]{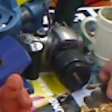}
	\caption{\label{fig:experiments_illustration}Illustration of some of the experiments. Left: we regress the joints orientation of a body skeleton given the 3D location of its joints. Center and right: synthetic and real image crops used in object pose estimation experiment.}
\end{figure}

Experiments were implemented using PyTorch 1.6~\cite{paszke_pytorch_2019}.
We illustrate some of the experiments performed in figure~\ref{fig:experiments_illustration}.

\subsection{Hue regression}
The hue regression example is implemented using a multi-layer perceptron with two hidden layers of 128 neurons. We use ReLU activations.
Training and test sets consist respectively of 200/20,000 random color samples with uniformly distributed hue and saturation, and a constant lightness of 0.5 (1.0 corresponding to white).
Training is performed using Adam optimizer with a learning rate of $10^{-3}$ for 5000 epochs on the whole training batch, using the mean cosine-similarity between predicted and target hue directions as training loss.
We do not display the actual hues in the paper's figure but instead use the \emph{twilight} colormap of Matplotlib to improve contrast for color-bind people and for grayscale printing.

\subsection{Point cloud alignment}
We follow closely the protocol of \cite{zhou_continuity_2019}.
Rotations are sampled uniformly according to the usual $SO(3)$ metric both for training and testing.
Sizes of training, validation and test sets are respectively 1890, 400, 400.
We train the network with a batch size of 10 for $2.6\cdot10^6$ iterations, using Adam optimizer~\cite{kingma_adam_2015} with a learning rate of $10^{-4}$. 100 rotations are considered for each point cloud during test.
We train each variant with 5 different random weight initializations.

\subsection{Inverse kinematics}
We use the provided test set (73 sequences, 107,119 poses), but split randomly the training sequences into a validation (73 sequences, 116,427 poses) and a training set (721 sequences, 1,030,236 poses). We train the network using Adam optimizer with a learning rate of $10^{-5}$ and a batch size of 64 for 1,960k iterations.

Following the reference implementation of Zhou~\etal~\cite{zhou_continuity_2019}, hyperparameters $\alpha_i$  introduced in equation~(10) of the paper are defined equal to $1/(3n)$ for all keypoints but 3 corresponding to the hips and for which $\alpha_i = 1/(3n) + 10/9$, where $n$ denotes the number of keypoints.
We train each variant with 5 different random weight initializations.

\subsection{Camera localization}
We choose $\delta = \gamma/8$ to weight the losses $\mathcal{L}_{Frobenius}$ and $\mathcal{L}_{quaternion}$ similarly, based on the following identities relative to the angle $\alpha$ between two rotations~\cite{bregier_defining_2018}:
\begin{equation}
\left\lbrace
\begin{aligned}
\min \| \bm{q} \pm \bm{q}^* \|^2 &= 4 \sin^2(\alpha/4) \underset{\alpha \rightarrow 0}{\sim} \alpha^2/4 \\
\| \bm{R} - \bm{R}^* \|^2_F &= 8 \sin^2(\alpha/2) \underset{\alpha \rightarrow 0}{\sim} 2 \alpha^2.
\end{aligned}
\right.
\end{equation}

Values for the hyperparameter $\beta$ are listed in table~\ref{tab:posenet_hyperparameters} and correspond to the ones suggested in the PoseNet implementation of Walch \etal~\cite{walch_image-based_2017}.
In accordance with their original work, we do not report results for the \emph{Street} and \emph{Great Court} sets, as training does not converge with the implementation provided.

\begin{table}
	\centering
	\caption{\label{tab:posenet_hyperparameters}Value of PoseNet hyperparameter $\beta$ for the different datasets.}
	{
		\small
		\begin{tabular}{cc}
			\hline
			King's College & 500 \\
			OldHospital & 1500 \\
			Shop Facade & 100 \\
			St Mary's Church & 250 \\
			\hline
		\end{tabular}
	}
\end{table}

\subsection{Object pose estimation}

We predict the orientation of an object from a $112 \times 112$ image crop centered on its centroid (see fig.~\ref{fig:experiments_illustration} right). We randomly select for each object 85\% of the crops for testing and 15\% for validation. We ignore symmetric objects which are out of the scope of this work (the reader is referred to~\cite{saxena_learning_2009, bregier_defining_2018, pitteri_object_2019}).
We perform a basic hyperparameter tuning by running for each variant two trainings of 30 epochs with learning rates of 0.1 and 1.0 (respectively of 0.1 and 0.01 for \emph{Matrix/Procrustes} and \emph{Matrix/Gram-Schmidt}) using SGD optimizer, and perform validation after each epoch to select the version to evaluate.

\section{\label{sec:proof}Proofs of mathematical statements}

We sketch in this section proofs of mathematical statements contained in the paper, especially regarding:
\begin{itemize}
	\item The existence of a mapping with connected pre-images (section~\ref{sec:projection_embedding}).
	\item the rotation vector representation (section~\ref{app:rotvec}).
	\item Procrustes mapping (section~\ref{app:procrustes}).
	\item Gram-Schmidt mapping (section~\ref{app:gramschmidt}).
	\item $4 \times 4$ symmetric matrix representation (section~\ref{app:symmatrix}).
	\item the restriction of rotation vector representation to small angles (section~\ref{app:smallrotvec}).
\end{itemize}

\paragraph{Remark regarding boundaries}
In this work, we do not consider the case of a manifold with boundaries or corners. It is indeed not a manifold in the usual sense -- \ie a topological space locally homeomorphic to $\mathbb{R}^d$ -- and is therefore out of our scope. A usual trick to deal with such space consists in restricting numerical applications to the manifold of interior points, as in our softmax
example where borders with exact 0/1 probabilities are ignored.

\subsection{\label{sec:projection_embedding}Existence of a mapping a with connected pre-images}
	
Let $Y$ be the smooth Riemannian manifold on which we want to perform regression.
The Nash embedding theorem~\cite{nash_c_1954} states that $Y$ can be isometrically embedded into some Euclidean space $E$.
Without loss of generality, we therefore consider $Y$ to be embedded in $E$.

Let $f$ be the closest point projection operator 
\begin{equation}
f: x \in X \rightarrow \arg\min_{y \in Y} \dist(x,y),
\end{equation}
where $X \subset E$ is the input domain on which this function is well-defined.
In this section, we show that the mapping $f$ satisfies the property of pre-images connectivity.

\begin{figure}
	\includegraphics{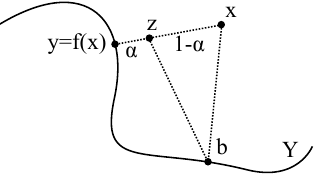}
	\caption{\label{fig:connectivity_notations}Notations used in section~\ref{sec:projection_embedding}.}
\end{figure}

Let $y \in Y$ and $x \in f^{-1}(y)$ an element of its pre-image by $f$.
We want to show that any point $z$ on the line segment $[x,y]$ is included in $f^{-1}(y)$,
\ie that for any $b \in Y \setminus \lbrace y \rbrace$,
$\dist(z, y) < \dist(z, b)$.
It is then straightforward to conclude the connectivity of $f^{-1}(y)$ from this result.
Figure~\ref{fig:connectivity_notations} illustrates the notations used.

Let $z$ be a point on $[x,y]$, $b \in Y \setminus \lbrace y \rbrace$ and $\alpha \in [0,1]$ be defined such that  $z=(1-\alpha) x + \alpha y$.
Triangle inequality ensures that
\begin{equation}
\dist(x, b) \leq \dist(x, z) + \dist(z, b),
\end{equation}
i.e. that
\begin{equation}
\dist(x, b) \leq (1-\alpha) \dist(x,y) + \dist(z, b)
\end{equation}
since $d(x,z)=(1-\alpha) \dist(x,y)$.

By definition of $f$, $\dist(x,y) < \dist(x,b)$ therefore
\begin{equation}
\dist(x, b) < (1-\alpha) \dist(x,b) + \dist(z, b)
\end{equation}
i.e.
\begin{equation}
\label{eq:ineq_connectivity}
\alpha \dist(x, b) < \dist(z, b).
\end{equation}

Combining this last inequality with the relations $d(z,y) = \alpha d(x,y)$ and $\dist(x,y) < \dist(x,b)$, we obtain our final result:
\begin{equation}
\begin{aligned}
d(z,y) &= \alpha \dist(x, y) \\
& < \alpha \dist(x, b) \\
& < \dist(z, b).
\end{aligned}
\end{equation}

\subsection{\label{sec:lemma_restriction}Useful lemma}
Let $f$ be a differentiable function between two differentiable manifolds $X$ and $Y$, and let $d$ be the dimension  of $Y$.

For $x \in X$, if there exists a submanifold $A \subset X$ containing $x$ for which
the restriction  $f_{\big|A}$ is a diffeomorphism from $A$ to $Y$,
then the Jacobian of $f$ is of rank $d$ at $x$ .

\paragraph{Proof}
Since $f_{\big|A}$ is a diffeomorphism to a $d$-dimensional manifold, its Jacobian is of rank $d$, therefore the rank of the Jacobian of $f$ at $x$ is at least $d$. But its rank is also at most $d$, since the image of $f$ is within $Y$.

\subsection{\label{app:rotvec} Rotation vector}

Rotation vector space corresponds to the usual Lie algebra associated to $SO(3)$.
As such, it is well known that the exponential mapping
\begin{equation}
f:\bm{x} \in \mathbb{R}^3 \rightarrow \exp(\bm{x}) \in SO(3)
\end{equation}
is surjective, continuous and differentiable.

The pre-image of a rotation parameterized by $\bm{x} \in \mathbb{R}^3$ consists in
\begin{equation}
\label{eq:rotvec_preimages}
f^{-1}(f(\bm{x})) = 
\left\lbrace
\begin{aligned}
\left\lbrace \left(1 + \cfrac{2\pi k}{\|\bm{x}\|} \right) \bm{x} \vert k \in \mathbb{Z} \right\rbrace &\text{ if } \|\bm{x}\| \neq 0. \\
\lbrace \bm{v} \in \mathbb{R}^3 \vert \|\bm{v}\| \in 2\pi \mathbb{N} \rbrace &\text{ else. }
\end{aligned}
\right.
\end{equation}
which is not connected.

In particular, for $\bm{x}$ of norm $2 \pi k$, with $k \in \mathbb{N}^*$, the pre-image of $f(\bm{x})$ restricted to a small neighborhood of $\bm{x}$ consists in a 2D-manifold (a subset of a sphere going through $\bm{x}$), therefore $f$ is not locally invertible at $\bm{x}$ and its Jacobian cannot therefore be of rank 3 (the dimension of $\mathbb{R}^3$ and $SO(3)$) at $\bm{x}$.
This result can also be obtained by observing an analytical expression of the derivatives of $f$ (replacing \eg $\bm{v}$ by $(2\pi, 0, 0)^\top$ in equation (III.1) of~\cite{gallego_compact_2015}, and observing the rank deficiency).

\paragraph{Restriction to a ball of limited angle.}
Given an arbitrary angle $\alpha \in ]0, \pi[$, a restriction of the exponential map from the open ball of radius $\alpha$ 
 $\left\lbrace \bm{x} \in \mathbb{R}^3 \vert \| \bm{x} \| < \alpha \right\rbrace$ to the set of rotations of angle strictly smaller than $\alpha$  is a diffeomorphism. It therefore satisfies properties of surjectivity, differentiability, full rank Jacobian and connected pre-images proposed in section 2 of the paper.

\subsection{\label{app:procrustes}Special Procrustes orthonormalization}
Given $d \in \mathbb{N}$ with $d \geq 2$, and a $d \times d$ matrix $\bm{M} \in \mathcal{M}_{d,d}(\mathbb{R})$, we consider the special orthogonal Procrustes problem consisting in finding the $d \times d$ rotation matrix $\bm{R} \in SO(d)$ closest to $\bm{M}$
\begin{equation}
\label{eq:procrustes_problem}
\bm{R} = \arg\min_{\bm{\tilde{R}} \in SO(3)} h(\bm{M}, \tilde{\bm{R}}),
\end{equation}
considering the square Frobenius distance
\begin{equation}
\label{eq:sqfrobnorm}
h(\bm{M}, \tilde{\bm{R}}) = \| \bm{\tilde{R}} - \bm{M} \|_F^2.
\end{equation}

Solutions to this problem consist in  $\bm{U} \bm{S} \bm{V}^\top$~\cite{schonemann_generalized_1966}\cite{umeyama_least-squares_1991},
where $\bm{U} \bm{D} \bm{V}^\top$ is a singular value decomposition of $\bm{M}$
such that $\bm{U}, \bm{V} \in O(d)$, $\bm{D} = \diag(d_1, \ldots, d_d)$ is a diagonal matrix verifying
\begin{equation}
d_1 \geq \ldots \geq d_d \geq 0,
\end{equation}
and such that $\bm{S} = \diag(1,\ldots,1,\det(\bm{U}) \det(\bm{V}))$.

The solution is unique if and only if $\det(\bm{M}) > 0$ or $d_{d-1} \neq d_{d}$.
We will denote $X$ the set of input matrices $\bm{M}$ satisfying this property
and $\Procrustes$ the function mapping any $\bm{M} \in X$ to the corresponding solution of the special Procrustes problem~\eqref{eq:procrustes_problem}.

\subsubsection{Surjectivity}

For any $\bm{R} \in SO(d)$, one can verify that $X$ contains $\bm{R}$
and that
$\Procrustes(\bm{R}) = \bm{R}$.
This proves the surjectivity of the $\Procrustes$ mapping.

\subsubsection{Differentiability}
Let $\bm{M}=(m_{ij})_{i,j \in \llbracket 1, d \rrbracket} \in X$ and $\bm{R} = \Procrustes(\bm{M})$.

Papadopoulo and Lourakis~\cite{papadopoulo_estimating_2000} showed how SVD decomposition could be locally differentiated,
and we  build upon their work to provide an analytical expression of the derivatives of $\Procrustes$ in the ambient space $\mathcal{M}_{d,d}(\mathbb{R})$.

Using notations introduced in the beginning of section~\ref{app:procrustes},
\begin{equation}
\bm{R} = \bm{U} \bm{S} \bm{V}^\top
\end{equation}
and therefore for any $i,j  \in \llbracket 1, d \rrbracket$,
\begin{equation}
\cfrac{\partial \bm{R}}{m_{ij}} = \cfrac{\partial \bm{U}}{m_{ij}} \bm{S} \bm{V}^\top 
+ \bm{U} \cfrac{\partial \bm{S}}{m_{ij}} \bm{V}^\top 
+ \bm{U} \bm{S} \left(\cfrac{\partial \bm{V}}{m_{ij}} \right)^\top.
\end{equation}

Since $\bm{S}$ is locally constant (being equal either to $\diag(1, \ldots, 1)$ or $\diag(1, \ldots, -1)$), its partial derivatives are null, therefore
\begin{equation}
\cfrac{\partial \bm{R}}{m_{ij}} = \cfrac{\partial \bm{U}}{m_{ij}} \bm{S} \bm{V}^\top + \bm{U} \bm{S} \left(\cfrac{\partial \bm{V}}{m_{ij}} \right)^\top.
\end{equation}

Since $\bm{U}, \bm{V} \in O(d)$, their partial derivatives have to encode this orthogonality constraint and can therefore be written 
\begin{equation}
\left\lbrace
\begin{aligned}
\cfrac{\partial \bm{U}}{m_{ij}} &= \bm{U} \bm{\Omega}_U^{ij} \\
\cfrac{\partial \bm{V}}{m_{ij}} &= -\bm{V} \bm{\Omega}_V^{ij},
\end{aligned}
\right.
\end{equation}
introducing $\bm{\Omega}_U^{ij}$ and $\bm{\Omega}_V^{ij}$ two skew-symmetric matrices (see \cite{papadopoulo_estimating_2000} for details).

Hence, the partial derivative can be rewritten
\begin{equation}
\label{eqref:procrustes_derivative_omega}
\cfrac{\partial \bm{R}}{m_{ij}} = \bm{U} \bm{\Omega}^{ij}  \bm{V}^\top
\end{equation}
with $\bm{\Omega}^{ij} = (\bm{\Omega}_U^{ij} \bm{S} + \bm{S} \bm{\Omega}_V^{ij})$.

Papadopoulo and Lourakis~\cite{papadopoulo_estimating_2000} showed that the elements of  $\bm{\Omega}_U^{ij}$ and $\bm{\Omega}_V^{ij}$ satisfy a particular set of linear equations.

Solving this system (while accounting for the potential flip of sign induced by $\bm{S}$) enables to provide an analytical expression for the elements of $\bm{\Omega}^{ij}$:
{
\footnotesize
\begin{equation}
\bm{\Omega}^{ij}_{kl} = 
\left\lbrace
\begin{aligned}
0 &\text{ if } k=l. \\
\cfrac{u_{ik} v_{jl}  - u_{il} v_{jk}}{d_k - d_l} &\text{ if } l=d, k \neq d \text{ and } \det(\bm{U}) \det(\bm{V}) < 0. \\
\cfrac{u_{ik} v_{jl}  - u_{il} v_{jk}}{d_l - d_k} &\text{ if } k=d, l \neq d \text{ and } \det(\bm{U}) \det(\bm{V}) < 0. \\
\cfrac{u_{ik} v_{jl} -u_{il} v_{jk}}{d_l+d_k} &\text{ else.}
\end{aligned}
\right.
\end{equation}
}
Since $\det(\bm{M}) = \det(\bm{U}) \det(\bm{D}) \det(\bm{V})$ and since $\det(\bm{D}) \geq 0$, we consider two different cases to show that the denominator does not cancel out in this expression:

If $\det(\bm{M}) > 0$, then $\det(\bm{U})\det(\bm{V}) > 0$ and $d_1 \geq \ldots \geq d_d > 0$, therefore elements of $\bm{\Omega}^{ij}$ are well-defined.

If $\det(\bm{M}) \leq 0$, then $\det(\bm{U})\det(\bm{V}) < 0$ and $d_d \neq d_{d-1}$ by definition of $X$. Therefore $d_1 \geq \ldots \geq d_{d-1} > d_d$ and elements of $\bm{\Omega}^{ij}$ are also well-defined.

We obtain a closed-form expression of the partial derivative of $\Procrustes$ by injecting this expression of $\bm{\Omega}^{ij}$ into equation~\eqref{eqref:procrustes_derivative_omega}.

\subsubsection{Full rank Jacobian}
Let $\bm{M} \in X$ and $\bm{R} = \Procrustes(\bm{M})$.
The function
\begin{equation}
g_{\bm{M}}: \tilde{\bm{R}} \in SO(d) \rightarrow \bm{M} \bm{R}^\top \tilde{\bm{R}}
\end{equation}
is differentiable and is the inverse of the restriction of $\Procrustes$ to $g_{\bm{M}}(SO(d))$,
\ie for any $\tilde{\bm{R}} \in SO(d)$,
\begin{equation}
\begin{aligned}
\Procrustes(g_{\bm{M}}(\tilde{\bm{R}})) &= \Procrustes(\bm{M}) \bm{R}^\top \tilde{\bm{R}} \\
& = \tilde{\bm{R}}.
\end{aligned}
\end{equation}
$\Procrustes$ is therefore a diffeomorphism from $g_{\bm{M}}(SO(d))$ to $SO(d)$ and its Jacobian is thus of rank $d$ (the dimension of $SO(d)$ manifold) according to lemma~\ref{sec:lemma_restriction}.

\subsubsection{Pre-images convexity}
Equation~\eqref{eq:sqfrobnorm} can be developed using a trace operator into
\begin{equation}
h(\bm{M}, \tilde{\bm{R}}) = \| \bm{M} \|_F^2 + \| \tilde{\bm{R}} \|_F^2 - 2 \trace( \bm{M}  \tilde{\bm{R}}^\top ).
\end{equation}
Since $\tilde{\bm{R}}$ is a rotation matrix, this expression is equivalent to
\begin{equation}
h(\bm{M}, \tilde{\bm{R}}) = \| \bm{M} \|_F^2 + \| \bm{I} \|_F^2 - 2 \trace( \bm{M} \tilde{\bm{R}}^\top ).
\end{equation}
where $\bm{I}$ denotes the $d\times d$ identity matrix.

Minimizing
\begin{equation}
\tilde{\bm{R}} \in SO(d) \rightarrow h(\bm{M}, \tilde{\bm{R}})
\end{equation} is therefore equivalent to maximizing
\begin{equation}
\label{eq:procrustes_trace}
\tilde{\bm{R}} \in SO(d) \rightarrow \trace( \bm{M}  \tilde{\bm{R}}^\top ).
\end{equation}

Let $\bm{R} \in SO(d)$, and $\bm{M}_1, \bm{M}_2 \in X$ two pre-images of $\bm{R}$ by $\Procrustes$, \ie
\begin{equation}
\Procrustes(\bm{M}_1) = \Procrustes(\bm{M}_2) = \bm{R}.
\end{equation}
By definition, $\bm{R}$ is the unique rotation maximizing the function defined  in equation~\eqref{eq:procrustes_trace}, considering $\bm{M}=\bm{M}_1$ and $\bm{M}=\bm{M}_2$.
For any $\alpha, \beta \in \mathbb{R}^{+*}$, thanks to the linearity of matrix multiplication and trace, 
$\bm{R}$ is therefore the unique maximizer of the function defined in equation~\eqref{eq:procrustes_trace} considering $\bm{M}=\alpha \bm{M}_1 + \beta \bm{M}_2$
\ie
\begin{equation}
\Procrustes(\alpha \bm{M}_1 + \beta \bm{M}_2) = \bm{R}.
\end{equation}
This proves in particular the convexity of the pre-image of $\bm{R}$ by $\Procrustes$.

\subsection{\label{app:gramschmidt}Special Gram-Schmidt orthonormalization}

\subsubsection{Gram-Schmidt process}
The Gram-Schmidt process enables to generate an orthonormal $n$-tuple of $d$-dimensional vectors $(\bm{e}_1, \ldots, \bm{e_n})$ from a linearly independent $n$-tuple of $d$-dimensional vectors $(\bm{v}_1, \ldots, \bm{v}_n)$ (assuming $n \leq d$). It is defined iteratively by $\bm{e}_1 = \bm{v}_1/\|\bm{v}_1\|$ and 
\begin{equation}
\label{eq:gramschmidt}
\bm{e}_k = \left(\bm{v}_k - \sum_{j=1}^{k-1} \proj_{\bm{e}_j}(\bm{v}_k) \right) / \left\| \bm{v}_k - \sum_{j=1}^{k-1} \proj_{\bm{e}_j}(\bm{v}_k) \right\|
\end{equation}
for $k=2 \ldots n$, where $\proj$ denotes the projection operator defined by
\begin{equation}
\proj_{\bm{a}}(\bm{b}) = \cfrac{\bm{a}^\top \bm{b}}{\| \bm{a} \|^2} \cdot \bm{a}
\end{equation}
for any pair of vectors $\bm{a}, \bm{b} \in \mathbb{R}^d$ with $\bm{a} \neq \bm{0}$.

\subsubsection{Special Gram-Schmidt orthonormalization}
Let $d \in \mathbb{N}$ with $d \geq 2$.
We define 
\begin{equation}
X = \lbrace \bm{V} \in \mathcal{M}_{d, d-1}(\mathbb{R}) \vert \rank(\bm{V}) = d-1 \rbrace
\end{equation}
the set of $d \times (d-1)$ matrices $\bm{V}$ whose column vectors $(\bm{v}_1, \ldots, \bm{v}_{d-1})$ are linearly independent.

We denote $\GramSchmidt$ the mapping from $X$ to $SO(d)$
defined for any $\bm{V}=(\bm{v}_1, \ldots, \bm{v}_{d-1}) \in X$ by
\begin{equation}
\GramSchmidt(\bm{V}) = \begin{pmatrix}
\bm{e}_1  & \ldots & \bm{e}_{d}
\end{pmatrix}
\end{equation}
where the $(\bm{e}_i)_{i=1,\ldots,d-1}$ are defined as in equation~\eqref{eq:gramschmidt}
and where $\bm{e}_d$ is chosen as the unique vector such that $\GramSchmidt(\bm{V}) \in SO(d)$.

 For $d=3$,  $\GramSchmidt$ corresponds to the \emph{6D} mapping of Zhou \etal~\cite{zhou_continuity_2019}, in which case $\bm{e}_3$ can be defined by $\bm{e}_3 = \bm{e}_1 \times \bm{e}_2$.

\subsubsection{Surjectivity}
For any $\bm{R} \in SO(d)$,
the matrix $(\bm{r}_1, \ldots, \bm{r}_{d-1})$ composed of the first $(d-1)$ columns of $\bm{R}$ belongs to $X$ and
\begin{equation}
\GramSchmidt((\bm{r}_1, \ldots, \bm{r}_{d-1})) = \bm{R}.
\end{equation}
This proves the surjectivity of $\GramSchmidt$.
\subsubsection{Differentiability}
All operations of the special Gram-Schmidt orthonormalization process are differentiable on $X$, and we can therefore deduce the differentiability of $\GramSchmidt$ on $X$. 

\subsubsection{Full rank Jacobian}
Let $\bm{V} \in X$ and $\bm{R} = \GramSchmidt(\bm{V})$.
The function
\begin{equation}
g_{\bm{V}}: \tilde{\bm{R}} \in SO(d) \rightarrow \tilde{\bm{R}} \bm{R}^{-1} \bm{V} \in X
\end{equation}
is differentiable and for any $\tilde{\bm{R}} \in SO(d)$,
\begin{equation}
\begin{aligned}
\GramSchmidt(g_{\bm{V}}(\tilde{\bm{R}})) &= \tilde{\bm{R}} \bm{R}^{-1} \GramSchmidt(\bm{V}) \\
&= \tilde{\bm{R}}.
\end{aligned}
\end{equation}
The restriction of $\GramSchmidt$ to  $g_{\bm{V}}(SO(d))$ is therefore a diffeomorphism from $g_{\bm{V}}(SO(d))$ to $SO(d)$. We conclude that the Jacobian of $\GramSchmidt$ is of rank $d$ (the dimension of $SO(d)$ manifold) at $\bm{V}$ thanks to lemma~\ref{sec:lemma_restriction}.

\subsubsection{Pre-images convexity}
Let $\bm{A}, \bm{B} \in X$ having the same image
\begin{equation}
\GramSchmidt(\bm{A}) = \GramSchmidt(\bm{B}) = \bm{E} \in SO(d).
\end{equation}
We denote respectively $(\bm{a}_1, \ldots, \bm{a}_{d-1})$, $(\bm{b}_1, \ldots, \bm{b}_{d-1})$ and $(\bm{e}_1, \ldots, \bm{e}_{d})$ the column vectors of $\bm{A}, \bm{B}$, and $\bm{E}$.

Let $\alpha, \beta \in \mathbb{R}^{+*}$ be two strictly positive scalars. We prove in this section that
\begin{equation}
\GramSchmidt(\alpha \bm{A} + \beta \bm{B}) = \bm{E}.
\end{equation}
It implies in particular that the pre-image of $\bm{E}$ by $\GramSchmidt$ is convex.

\paragraph{Proof}
Let $(\bm{c}_1, \ldots, \bm{c}_{d})$ be the column vectors of $\GramSchmidt(\alpha \bm{A} + \beta \bm{B})$.

By definition,
\begin{equation}
\bm{e}_1 = \cfrac{\bm{a}_1}{ \| \bm{a}_1\| } = \cfrac{\bm{b}_1}{ \| \bm{b}_1 \|}
\end{equation}
which implies that $\bm{a}_1$ and $\bm{b}_1$ are non-zero collinear vectors pointing in the same direction. $\alpha \bm{a}_1 + \beta \bm{b}_1$ is consequently also a non-null vector aligned with $\bm{e}_1$, therefore
\begin{equation}
\begin{aligned}
\bm{e}_1 &= \cfrac{\alpha \bm{a}_1 + \beta \bm{b}_1}{ \| \alpha \bm{a}_1 + \beta \bm{b}_1 \| } = \bm{c}_1.\\
\end{aligned}
\end{equation}

Let $k \in \llbracket 2, d-1 \rrbracket$ such that for all $j \in \llbracket 1, k-1 \rrbracket$, $\bm{e}_j = \bm{c}_j$.
Since $\proj$ is a linear operator, the following equalities hold true:
\begin{equation}
\begin{aligned}
&\alpha \bm{a}_k + \beta \bm{b}_k - \sum_{j=1}^{k-1} \proj_{\bm{c}_j}(\alpha \bm{a}_j + \beta \bm{b}_j) \\
&= \alpha \bm{a}_k + \beta \bm{b}_k - \sum_{j=1}^{k-1} \proj_{\bm{e}_j}(\alpha \bm{a}_j + \beta \bm{b}_j) \\
&= \alpha \underbrace{\left( \bm{a}_k - \sum_{j=1}^{k-1} \proj_{\bm{e}_j}(\bm{a}_j) \right)}_{\bm{f}} + \beta \underbrace{\left( \bm{b}_k - \sum_{j=1}^{k-1} \proj_{\bm{e}_j}(\bm{b}_j) \right)}_{\bm{g}}.
\end{aligned}
\end{equation}
From the definition of $\bm{e}_k$, we know that underbraced terms $\bm{f}$ and $\bm{g}$ consist in non-null vectors aligned with $\bm{e}_k$. $\alpha \bm{f} + \beta \bm{g}$ is therefore a non-null vector aligned with $\bm{e}_k$ as well
and consequently
\begin{equation}
\bm{c}_k = \cfrac{\alpha \bm{f} + \beta \bm{g}}{\| \alpha \bm{f} + \beta \bm{g} \|} = \bm{e}_k.
\end{equation}
This proves by induction that $\bm{e}_k = \bm{c}_k$ for any $k \in \llbracket 1, d-1 \rrbracket$.

Since $\bm{c}_d$ only depends on $(\bm{c}_k)_{k=1 \ldots d-1}$, we deduce that $\bm{c}_d = \bm{e}_d$, which concludes the proof.

\subsubsection{\label{app:lemma_weighted_procrustes}Lemma: weighted special Procrustes orthonormalization}
Let $\bm{M}, \bm{\Lambda} \in \mathcal{M}_{d,n}(\mathbb{R})$ be two arbitrary $d \times n$ matrices (considering an arbitrary $n \in \mathbb{N}^{*})$.
Minimizing the function
\begin{equation}
\bm{R} \in SO(d) \rightarrow \| \bm{R} \bm{\Lambda} - \bm{M} \|_F^2
\end{equation}
is equivalent to minimizing
\begin{equation}
\bm{R} \in SO(d) \rightarrow \| \bm{R} - \bm{M} \bm{\Lambda}^\top\|_F^2.
\end{equation}

\paragraph{Proof}
For any $\bm{R} \in SO(d)$,
\begin{equation}
\| \bm{R} \bm{\Lambda} - \bm{M} \|_F^2 = \| \bm{\Lambda} \|_F^2 + \| \bm{M} \|_F^2 - 2 \trace(\bm{R} \bm{\Lambda} \bm{M}^\top)
\end{equation}
and
\begin{equation}
\| \bm{R}  - \bm{M} \bm{\Lambda}^\top \|_F^2 = \| \bm{I} \|_F^2 + \| \bm{M} \bm{\Lambda}^\top \|_F^2 - 2 \trace(\bm{R} \bm{\Lambda} \bm{M}^\top).
\end{equation}
Both expressions are equal with respect to $\bm{R}$ up to a constant. Minimizing one with respect to $\bm{R}$ is therefore equivalent to minimizing the other.

\subsubsection{Link between Gram-Schmidt and Procrustes}
We denote $(\bm{\Lambda}_{n})_{n \in \mathbb{N}}$ a series of $d \times (d-1)$ matrices
\begin{equation}
\bm{\Lambda}_{n} = \begin{pmatrix}
\lambda_{1,n} & 0 & 0 \\
0 & \ddots & 0 \\
0 & 0 & \lambda_{d-1,n} \\
0 & 0 & 0\\
\end{pmatrix}
\end{equation}
satisfying for any $n \in \mathbb{N}$
\begin{equation}
\left\lbrace
\begin{aligned}
&\lambda_{1,n} = 1 \\
&\lambda_{i,n} > 0 \text{ for } i=1,\ldots, d-1 \\
\end{aligned}
\right.
\end{equation}
and defined such that $\lambda_{i,n} = o(\lambda_{i-1,n}) \text{ for } i \in \llbracket 2, d-1 \rrbracket$, using Landau notations.

For $\bm{M} \in X$, we show in this section that if the series $(\Procrustes(\bm{M}\bm{\Lambda}_n^\top))_{n \in \mathbb{N}}$
converges, its limit consists in $\GramSchmidt(\bm{M})$.

\paragraph{Proof}
Let $n \in \mathbb{N}$, $\bm{M} \in X$ and $\bm{R} \in SO(d)$.
We denote $(\bm{r}_1, \ldots, \bm{r}_d)$ and $(\bm{m}_1, \ldots, \bm{m}_{d-1})$ the 
respective column vectors of $\bm{R}$ and $\bm{M}$.
We denote
\begin{equation}
E(\bm{R}, \bm{\Lambda}_n) = \|\bm{R} \bm{\Lambda}_n - \bm{M} \|_F^2
\end{equation}
the minimization problem to which $\Procrustes(\bm{M}\bm{\Lambda}_n^\top)$ is associated (see lemma~\ref{app:lemma_weighted_procrustes}).
$\Procrustes(\bm{M}\bm{\Lambda}_n^\top)$ is well defined because $\rank(\bm{M}\bm{\Lambda}_n^\top) = d-1$, therefore the smallest singular value of $\bm{M}\bm{\Lambda}_n^\top$ is 0 with a multiplicity of 1.

It is straightforward to show that for any $j \in \llbracket 1,d-1 \rrbracket$,
{
\small
\begin{equation}
\begin{aligned}
\label{eq:proof_j}
E(\bm{R}, \bm{\Lambda}_n) &= \|\bm{M}\|^2 + \sum_{i=1}^{d-1} \lambda_{i,n}^2 - 2 \sum_{i=1}^{d-1} \lambda_{i,n} \bm{r}_i^\top \bm{m}_i \\
&=  \|\bm{M}\|^2 + \sum_{i=1}^{j} \lambda_{i,n}^2 - 2 \sum_{i=1}^{j} \lambda_{i,n} \bm{r}_i^\top \bm{m}_i + o(\lambda_{j,n}).
\end{aligned}
\end{equation}
}
In particular, for $j=1$,
\begin{equation}
\label{eq:proof_j=1}
E(\bm{R}, \bm{\Lambda}_n) =  \|\bm{M}\|^2 + 1 - 2  \bm{r}_1^\top \bm{m}_1 + o(1).
\end{equation}

Assuming that $(\Procrustes(\bm{M}\bm{\Lambda}_n^\top))_{n \in \mathbb{N}}$ admits a limit $\bm{R}^* = (\bm{r}^*_1 \ldots \bm{r}^*_d) \in SO(d)$, this limit should tend to minimize equation~\eqref{eq:proof_j=1}, \ie maximize $\bm{r}_1^\top \bm{m}_1$. Therefore $\bm{r}^*_1$ should be equal to $\bm{m}_1/ \|\bm{m}_1\|$.

Similarly, assuming that the expressions of $\bm{r}^*_1, \ldots ,\bm{r}^*_{j-1}$ are known for a given $j \in \llbracket 1, d-1 \rrbracket$,
$\bm{R}^*$ should tend to minimize equation~\eqref{eq:proof_j}, \ie tend to maximize $\bm{r}_j^\top \bm{m}_j$, which is obtained for
\begin{equation}
\bm{r}^*_j = \left( \bm{m}_j - \sum_{i=1}^{j-1} \proj_{\bm{r}^*_i}(\bm{m}_j) \right) / \left\| \bm{m}_j - \sum_{i=1}^{j-1} \proj_{\bm{r}^*_i}(\bm{m}_j) \right\|.
\end{equation}

A limit of $(\Procrustes(\bm{M}\bm{\Lambda}_n^\top))_{n \in \mathbb{N}}$ should therefore correspond to the special Gram-Schmidt orthonormalization of $\bm{M}$.

\subsection{\label{app:symmatrix}Symmetric matrix}
For $\bm{a} = (a_1, \ldots a_{10})^\top \in \mathbb{R}^{10}$,
we denote $\bm{S}_{\bm{a}}$ the $4 \times 4$ symmetric matrix
\begin{equation}
\bm{S}_{\bm{a}} =
\begin{pmatrix}
a_1 & a_2 & a_3 & a_4 \\
a_2 & a_5 & a_6 & a_7 \\
a_3 & a_6 & a_8 & a_9 \\
a_4 & a_7 & a_9 & a_{10} \\ 
\end{pmatrix}.
\end{equation}

Let $\bm{V} \diag(\lambda_1, \ldots, \lambda_4) \bm{V}^\top$ an eigenvalue decomposition of $\bm{S}_{\bm{a}}$, with $\bm{V} \in O(4)$ the matrix of eigenvectors and $\lambda_1, \ldots, \lambda_4$ the corresponding eigenvalues, sorted in ascending order.

The optimization problem proposed by Peretroukhin \etal~\cite{peretroukhin_smooth_2020}
\begin{equation}
\label{eq:symmatrix}
\min_{\bm{q} \in \mathbb{R}^4 / \| \bm{q} \| = 1} \bm{q}^\top \bm{S}_{\bm{a}} \bm{q}
\end{equation}
admits a unique pair of antipodal solutions $\pm \bm{q}$ if and only if $\lambda_1 \neq \lambda_2$, and this solution $\bm{q}$ consists in the first column vector of $\bm{V}$, corresponding to the eigenvector associated with the smallest eigenvalue $\lambda_1$.

Let $X \subset \mathbb{R}^{10}$ be the set of vectors $\bm{a}$ for which a minimizer $\bm{q}$ of~\eqref{eq:symmatrix} is unique (up to sign), and let $\symmatrix:X \rightarrow \mathbb{RP}^3$ denote the function mapping $\bm{a}$ to $\bm{q}$.

Peretroukhin \etal~\cite{peretroukhin_smooth_2020} showed that this mapping was surjective and differentiable, and proposed an expression of its Jacobian.
We show in the next subsections how it satisfies properties regarding the rank of its Jacobian and the connectivity of its pre-images.

\subsubsection{Full rank Jacobian}
Let $\bm{a} \in X$ and $\bm{q} = \symmatrix(x)$ a corresponding unit quaternion.

For any unit quaternion $\tilde{\bm{q}} = w + x i + y j + z k  \in \mathbb{RP}^3$, we denote
\begin{equation}
\bm{M}_{\tilde{\bm{q}}} = \begin{pmatrix}
w & -x & - y & -z\\
x & w & -z & y \\
y & z & w & -x \\
z & -y & x & w
\end{pmatrix} \in O(4)
\end{equation}
a matrix representation of $\tilde{\bm{q}}$, and we define the function
\begin{equation}
g_{\bm{x}}: \tilde{\bm{q}} \in \mathbb{RP}^3 \rightarrow \bm{M}_{\tilde{\bm{q}}^{-1}}^\top \bm{M}_{\bm{q}}^\top \bm{S}_{\bm{a}} \bm{M}_{\bm{q}} \bm{M}_{\tilde{\bm{q}}^{-1}}.
\end{equation}
$g_{\bm{x}}$ is well-defined, as $g_{\bm{x}}(\tilde{\bm{q}}) = g_{\bm{x}}(-\tilde{\bm{q}})$ for any $\tilde{\bm{q}} \in \mathbb{RP}^3$.
It returns symmetric matrices whose eigenvalues are identical to those of $\bm{S}_{\bm{a}}$,
and for any unit quaternions $\tilde{\bm{q}}, \bm{q}^*$,
\begin{equation}
\begin{aligned}
&\pm \bm{q}^* = \symmatrix(g_{\bm{x}}(\tilde{\bm{q}})) \\
&\Leftrightarrow  \pm \bm{M}_{\bm{q}} \bm{M}_{\tilde{\bm{q}}^{-1}} \bm{q}^* = \symmatrix(\bm{S}_{\bm{a}}) \\
&\Leftrightarrow  \pm \bm{q}\tilde{\bm{q}}^{-1} \bm{q}^* = \pm \bm{q} \\
&\Leftrightarrow  \pm \bm{q}^* = \pm \tilde{\bm{q}}.
\end{aligned}
\end{equation}
Identifying a symmetric matrix with its vectorized entries, the range of $g_{\bm{x}}$ is therefore included in $X$, and $g_{\bm{x}}$ is the differentiable inverse of the restriction of $\symmatrix$ to $g_{\bm{x}}(\mathbb{RP}^3)$.
We conclude that the Jacobian of $\symmatrix$ is of rank 3 (the dimension of $\mathbb{RP}^3$ manifold)  at $\bm{a}$ using lemma~\ref{sec:lemma_restriction}.

\subsubsection{Pre-images convexity}
Let $\bm{a}, \bm{b} \in X$ and $ \pm \bm{q^*} \in \mathbb{RP}^3$ such that
\begin{equation}
\symmatrix(\bm{a}) = \symmatrix(\bm{b}) = \pm \bm{q^*}.
\end{equation}
For any $\alpha, \beta \in \mathbb{R}^{+*}$, both functions
\begin{equation}
\bm{q} \in \mathbb{R}^4 / \|\bm{q} \| = 1 \rightarrow \alpha \bm{q}^\top
 \bm{S}_{\bm{a}} \bm{q}
\end{equation} 
  and 
\begin{equation}
\bm{q} \in \mathbb{R}^4 / \|\bm{q} \| = 1 \rightarrow \beta \bm{q}^\top
\bm{S}_{\bm{b}} \bm{q}
\end{equation}
admit a unique pair of minimizers $\pm \bm{q^*}$.
$\pm \bm{q^*}$ are therefore also the only minimizers of 
\begin{equation}
\bm{q} \in \mathbb{R}^4 / \|\bm{q} \| = 1 \rightarrow  \bm{q}^\top
(\alpha \bm{S}_{\bm{a}} + \beta \bm{S}_{\bm{b}}) \bm{q}.
\end{equation}
\ie of
\begin{equation}
\bm{q} \in \mathbb{R}^4 / \|\bm{q} \| = 1 \rightarrow  \bm{q}^\top
\bm{S}_{(\alpha \bm{a} + \beta \bm{b})} \bm{q}.
\end{equation}
We deduce that $\alpha \bm{a} + \beta \bm{b} \in X$ and that $\symmatrix(\alpha \bm{a} + \beta \bm{b}) = \pm \bm{q}^*$.
In particular, it implies that the pre-image set $\symmatrix^{-1}(\pm \bm{q^*})$ is convex.

\subsection{\label{app:smallrotvec}Limited range rotation vector}
Given $n \in \mathbb{N}^*$ and $\alpha \in \mathbb{R}^{+*}$, we denote $B_{\alpha}$ the open ball of $\mathbb{R}^n$ of radius $\alpha$.

The function
\begin{equation}
g_{\alpha}: \bm{x} \in \mathbb{R}^n \rightarrow  
\left\lbrace
\begin{aligned}
&\alpha \cfrac{\tanh(\|\bm{x}\|)}{ \bm{\|x\|}} \bm{x} \in B_{\alpha} \text{ if } \bm{x} \neq \bm{0}, \\
&\bm{0} \text{ else}
\end{aligned}
\right.
\end{equation}
 is continuous and differentiable everywhere, with partial derivatives where $\bm{x} \neq \bm{0}$ equal to
 {
\footnotesize
\begin{equation}
\left\lbrace
\begin{aligned}
\cfrac{\partial g_{\alpha,i}}{\partial x_i} &= \alpha \left( -\cfrac{x_i^2 \tanh(\|x\|)}{\|\bm{x}\|^3} + \cfrac{\tanh(\|\bm{x}\|)}{\|\bm{x}\|} + \cfrac{x_i^2 \sech^2(\|\bm{x}\|)}{\|\bm{x}\|^2} \right) \\
\cfrac{\partial g_{\alpha,i}}{\partial x_j} &= \alpha \left( \cfrac{x_i x_j \sech^2(\|x\|)}{\|x\|^2} - \cfrac{x_i x_j \tanh(\|\bm{x}\|)}{\|\bm{x}\|^3} \right) \text{ for } i \neq j.
\end{aligned}
\right.
\end{equation}
}
Taylor series expansion shows the continuity of $g_{\alpha}$ and its derivatives in $\bm{0}$, which are locally equivalent to
\begin{equation}
\left\lbrace
\begin{aligned}
g_{\alpha}(\bm{x}) &\sim \alpha \bm{x} \rightarrow \bm{0}\\
\cfrac{\partial g_{\alpha,i}}{\partial x_i} &\sim \alpha \left( 1 - \cfrac{5}{6} x_i^2 \right) \rightarrow \alpha\\
\cfrac{\partial g_{\alpha,i}}{\partial x_j} &\sim -\cfrac{2}{3} \alpha  x_i x_j \rightarrow 0  \text{ for } i \neq j.
\end{aligned}
\right.
\end{equation}

$g_{\alpha}$ admits moreover a differentiable inverse
\begin{equation}
g_{\alpha}^{-1}: \bm{y} \in B_{\alpha} \rightarrow 
\left\lbrace
\begin{aligned}
&\cfrac{\tanh^{-1}(  \| \bm{y} \| / \alpha)}{ \| \bm{y} \| } \bm{y} \text{ if } \bm{y} \neq \bm{0}, \\
&\bm{0} \text{ else},
\end{aligned}
\right.
\end{equation}
and is therefore a diffeomorphism from $\mathbb{R}^n$ to $B_{\alpha}$.

Considering $n=3$ and $\alpha \in ]0, \pi[$, the exponential map $\exp$ is a diffeomorphism from $B_{\alpha}$ to the manifold of 3D rotations of angle strictly smaller than $\alpha$. The composed function $g_{\alpha} \circ exp$ is therefore a diffeomorphism from $\mathbb{R}^n$ to this set of rotations, and it therefore satisfies properties of surjectivity, differentiability, full rank Jacobian and pre-image connectivity proposed in section 2 of the paper.


{\small
	\bibliographystyle{ieee_fullname}
	\bibliography{biblio}
}

%% file: figures/toy_example_architecture.pdf_tex
\begingroup%
  \makeatletter%
  \providecommand\color[2][]{%
    \errmessage{(Inkscape) Color is used for the text in Inkscape, but the package 'color.sty' is not loaded}%
    \renewcommand\color[2][]{}%
  }%
  \providecommand\transparent[1]{%
    \errmessage{(Inkscape) Transparency is used (non-zero) for the text in Inkscape, but the package 'transparent.sty' is not loaded}%
    \renewcommand\transparent[1]{}%
  }%
  \providecommand\rotatebox[2]{#2}%
  \newcommand*\fsize{\dimexpr\f@size pt\relax}%
  \newcommand*\lineheight[1]{\fontsize{\fsize}{#1\fsize}\selectfont}%
  \ifx\svgwidth\undefined%
    \setlength{\unitlength}{56.69291339bp}%
    \ifx\svgscale\undefined%
      \relax%
    \else%
      \setlength{\unitlength}{\unitlength * \real{\svgscale}}%
    \fi%
  \else%
    \setlength{\unitlength}{\svgwidth}%
  \fi%
  \global\let\svgwidth\undefined%
  \global\let\svgscale\undefined%
  \makeatother%
  \begin{picture}(1,2.1)%
    \lineheight{1}%
    \setlength\tabcolsep{0pt}%
    \put(0,0){\includegraphics[width=\unitlength,page=1]{toy_example_architecture.pdf}}%
    \put(0.42302735,1.90894218){\color[rgb]{0,0,0}\makebox(0,0)[t]{\lineheight{1.25}\smash{\begin{tabular}[t]{c}RGB value $a$\end{tabular}}}}%
    \put(0,0){\includegraphics[width=\unitlength,page=2]{toy_example_architecture.pdf}}%
    \put(0.42383254,1.29752632){\color[rgb]{0,0,0}\makebox(0,0)[t]{\lineheight{1.25}\smash{\begin{tabular}[t]{c}MLP $h$\end{tabular}}}}%
    \put(0.73195517,1.01027504){\color[rgb]{0,0,0}\makebox(0,0)[t]{\lineheight{1.25}\smash{\begin{tabular}[t]{c}\parbox{2cm}{\centering$ x=h(a)$ \\ $\in X$}\end{tabular}}}}%
    \put(0.42174819,0.04187523){\color[rgb]{0,0,0}\makebox(0,0)[t]{\lineheight{1.25}\smash{\begin{tabular}[t]{c}$y=f(x) \in Y$\end{tabular}}}}%
    \put(0,0){\includegraphics[width=\unitlength,page=3]{toy_example_architecture.pdf}}%
    \put(0.42239802,0.59435097){\color[rgb]{0,0,0}\makebox(0,0)[t]{\lineheight{1.25}\smash{\begin{tabular}[t]{c}Mapping $f$\end{tabular}}}}%
    \put(0,0){\includegraphics[width=\unitlength,page=4]{toy_example_architecture.pdf}}%
  \end{picture}%
\endgroup%

%% file: merged_posenet_table.tex
\resizebox{1.0\textwidth}{!}{%
\begin{tabular}{|c|ccccc|c|cccc|}
\hline
Loss              & $\mathcal{L}_{original}$                                                 & \multicolumn{5}{|c|}{$\mathcal{L}_{quaternion}$ (loss~\eqref{eq:posenet_quaternion_loss})}            & \multicolumn{4}{|c|}{$\mathcal{L}_{Frobenius}$ (loss~\eqref{eq:posenet_frobenius_loss})}                                                                                                                                                                                          \\
\hline
Representation    & \multicolumn{1}{c|}{quaternion}                            & quaternion                                         & Rot-vec                                          & 6D                                       & Procrustes                                        & Rot-vec 180\textdegree               & quaternion                                         & Rot-vec                                          & 6D                                       & Procrustes              \\
\hline                                                                                                                                                                                                                                                                              
KingsCollege & 1.54m, \textbf{4.25\textdegree{}} & 1.82m, 4.73\textdegree{} & 1.64m, 4.77\textdegree{} & 1.64m, 4.82\textdegree{} & \textbf{1.46m}, 5.15\textdegree{} & 1.51m, 138.01\textdegree{}      & 1.87m, 4.98\textdegree{} & \textbf{1.57m}, 4.87\textdegree{} & 1.70m, \textbf{4.56\textdegree{}} & 1.87m, 5.17\textdegree{} \\
OldHospital & 2.64m, 5.20\textdegree{} & 2.36m, 5.15\textdegree{} & \textbf{1.95m}, 5.88\textdegree{} & 2.52m, 4.90\textdegree{} & 2.23m, \textbf{4.63\textdegree{}} & 2.32m, 155.73\textdegree{}       & 2.53m, 5.23\textdegree{} & 2.56m, 4.97\textdegree{} & 2.68m, 4.83\textdegree{} & \textbf{2.37m}, \textbf{4.39\textdegree{}} \\
ShopFacade & \textbf{2.04m}, 10.07\textdegree{} & 2.21m, 10.49\textdegree{} & 1.44m, 9.68\textdegree{} & 2.40m, 10.61\textdegree{} & 1.57m, \textbf{8.25\textdegree{}} & 1.55m, 143.32\textdegree{}     & 2.16m, 9.56\textdegree{} & \textbf{1.27m}, \textbf{7.55\textdegree{}} & 1.90m, 9.22\textdegree{} & 1.50m, 8.75\textdegree{} \\
StMarysChurch & \textbf{2.06m, 7.45\textdegree{}} & 2.21m, 10.13\textdegree{} & 1.94m, 7.88\textdegree{} & 2.22m, 8.17\textdegree{} & 2.11m, 8.08\textdegree{} & 2.02m, 165.23\textdegree{}             & \textbf{1.97m}, 8.32\textdegree{} & 2.02m, 7.82\textdegree{} & 2.14m, 8.01\textdegree{} & 2.01m, \textbf{7.28\textdegree{}} \\
\hline
Mean  & 2.07m, 6.74\textdegree{} & 2.15m, 7.63\textdegree{} & \textbf{1.74m}, 7.05\textdegree{} & 2.19m, 7.13\textdegree{} & 1.84m, \textbf{6.53\textdegree{}} & 1.85m, 150.57\textdegree{}             & 2.13m, 7.02\textdegree{} & \textbf{1.85m, 6.30\textdegree{}} & 2.10m, 6.65\textdegree{} & 1.94m, 6.40\textdegree{}\\
\hline
\end{tabular}%
}

%% file: object_table.tex
\begin{tabular}{|c|cccc|cc|}

\hline
Object & Quat. & Rot-vec & 6D & Procr. & Mat/GS. & Mat/Procr. \\
\hline
ape & 0.145 & 0.119 & 0.085 & \textbf{0.080} &  0.109 & \textbf{0.107} \\
bench vise & 0.062 & 0.068 & 0.049 & \textbf{0.042} & \textbf{0.067} & 0.069 \\
camera & 0.139 & 0.145 & 0.084 & \textbf{0.067} & 0.236 & \textbf{0.130} \\
watering can & 0.162 & 0.169 & 0.134 & \textbf{0.132} & 0.242 & \textbf{0.239} \\
cat & 0.078 & 0.091 & \textbf{0.048} & 0.049 & 0.174 & \textbf{0.111} \\
cup & \textbf{0.215} & 0.237 & 0.219 & 0.221 & 0.260 & \textbf{0.232} \\
driller & 0.106 & 0.089 & 0.059 & \textbf{0.048} & 0.186 & \textbf{0.144} \\
duck & 0.100 & 0.118 & \textbf{0.050} & 0.051 & 0.135 & \textbf{0.129} \\
hole puncher & 0.177 & 0.165 & 0.137 & \textbf{0.126} & 0.218 & \textbf{0.176} \\
iron & 0.051 & 0.053 & 0.041 & \textbf{0.037} & \textbf{0.055} & 0.079 \\
lamp & 0.061 & 0.054 & \textbf{0.032} & 0.035 & 0.088 & \textbf{0.081} \\
phone & 0.101 & 0.129 & \textbf{0.073} & 0.077 & 0.155 & \textbf{0.122} \\
\hline
Mean & 0.116 & 0.120 & 0.084 & \textbf{0.080} & 0.160 & \textbf{0.135} \\
\hline
\end{tabular}